\documentclass[journal,twoside]{IEEEtran}

\ifCLASSINFOpdf
  \usepackage[pdftex]{graphicx}
  \graphicspath{{figures/}}
  \DeclareGraphicsExtensions{.pdf,.jpeg,.png}
\else
  \usepackage[dvips]{graphicx}
  \graphicspath{{figures/}}
  \DeclareGraphicsExtensions{.eps}
\fi

\usepackage{diagbox}
\usepackage{multirow}

\usepackage{cite}

\usepackage{amsmath}
\usepackage{amsfonts}

\usepackage{algorithm}
\usepackage{algorithmic}

\usepackage{array}

\ifCLASSOPTIONcompsoc
  \usepackage[caption=false,font=normalsize,labelfont=sf,textfont=sf]{subfig}
\else
  \usepackage[caption=false,font=footnotesize]{subfig}
\fi
\usepackage{float}

\usepackage{threeparttable}

\usepackage{lipsum}

\usepackage{url}
\usepackage[colorlinks,linkcolor=red,anchorcolor=blue,citecolor=green,CJKbookmarks=True]{hyperref}

\begin{document}
\title{Attention and Prediction Guided Motion Detection for Low-Contrast Small Moving Targets}

\author{Hongxin Wang, Jiannan Zhao, Huatian Wang, Cheng Hu, Jigen Peng, and Shigang Yue, \IEEEmembership{Senior Member,~IEEE} 
	\thanks{This work was supported in part by the National Natural Science Foundation of China under Grant 12031003, Grant 62103112, and Grant 11771347, in part by the European Union’s Horizon 2020 research and innovation programme under the Marie Sklodowska-Curie grant agreement No 691154 STEP2DYNA and No 778062 ULTRACEPT, in part by the China Postdoctoral Science Foundation under Grant 2021M700921 and Grant 2019M662837. \emph{(Corresponding authors:  Shigang Yue; Jigen Peng.)}}
	\thanks{Hongxin Wang, Cheng Hu, and Shigang Yue are with the Machine Life and Intelligence Research Center, Guangzhou University, Guangzhou 510006, China, and also with
		the Computational Intelligence Laboratory, School of Computer Science, University of Lincoln, Lincoln LN6 7TS, U.K. (email: howang@lincoln.ac.uk; syue@lincoln.ac.uk).}
	\thanks{Jiannan Zhao is with the School of Electrical Engineering, Guangxi University, Nanning 530004, China, and also with the Computational Intelligence Laboratory, School of Computer Science, University of Lincoln, Lincoln LN6 7TS, U.K.}
	\thanks{Huatian Wang is with the Northwest Institute of Mechanical and Electrical Engineering, Xianyang 712099, China, and also with the Computational Intelligence Laboratory, School of Computer Science, University of Lincoln, Lincoln LN6 7TS, U.K.}
	\thanks{Jigen Peng is with the School of Mathematics and Information Science, Guangzhou University, Guangzhou 510006, China (email: jgpeng@gzhu.edu.cn).}
}

\markboth{IEEE TRANSACTIONS ON CYBERNETICS}
{Wang \MakeLowercase{\textit{et al.}}: Attention and Prediction Guided Motion Detection}
%



\maketitle

\begin{abstract}
 Small target motion detection within complex natural environments is an extremely challenging task for autonomous robots. Surprisingly, the visual systems of insects have evolved to be highly efficient in detecting mates and tracking prey, even though targets occupy as small as a few degrees of their visual fields. The excellent sensitivity to small target motion relies on a class of specialized neurons called small target motion detectors (STMDs). However, existing STMD-based models are heavily dependent on visual contrast and perform poorly in complex natural environments where small targets generally exhibit extremely low contrast against neighbouring backgrounds. In this paper, we develop an attention and prediction guided visual system to overcome this limitation. The developed visual system comprises three main subsystems, namely, an attention module, an STMD-based neural network, and a prediction module. The attention module searches for potential small targets in the predicted areas of the input image and enhances their contrast against complex background. The STMD-based neural network receives the contrast-enhanced image and discriminates small moving targets from background false positives. The prediction module foresees future positions of the detected targets and generates a prediction map for the attention module. The three subsystems are connected in a recurrent architecture allowing information to be processed sequentially to activate specific areas for small target detection. Extensive experiments on synthetic and real-world datasets demonstrate the effectiveness and superiority of the proposed visual system for detecting small, low-contrast moving targets against complex natural environments.
\end{abstract}







\begin{IEEEkeywords}
	Bioinspiration, small target motion detection, prediction, robotic visual perception, complex natural environment.
\end{IEEEkeywords}

%
\IEEEpeerreviewmaketitle

\section{Introduction}

\begin{figure}[t!]
	\centering
	\subfloat[]{\includegraphics[width=0.24\textwidth]{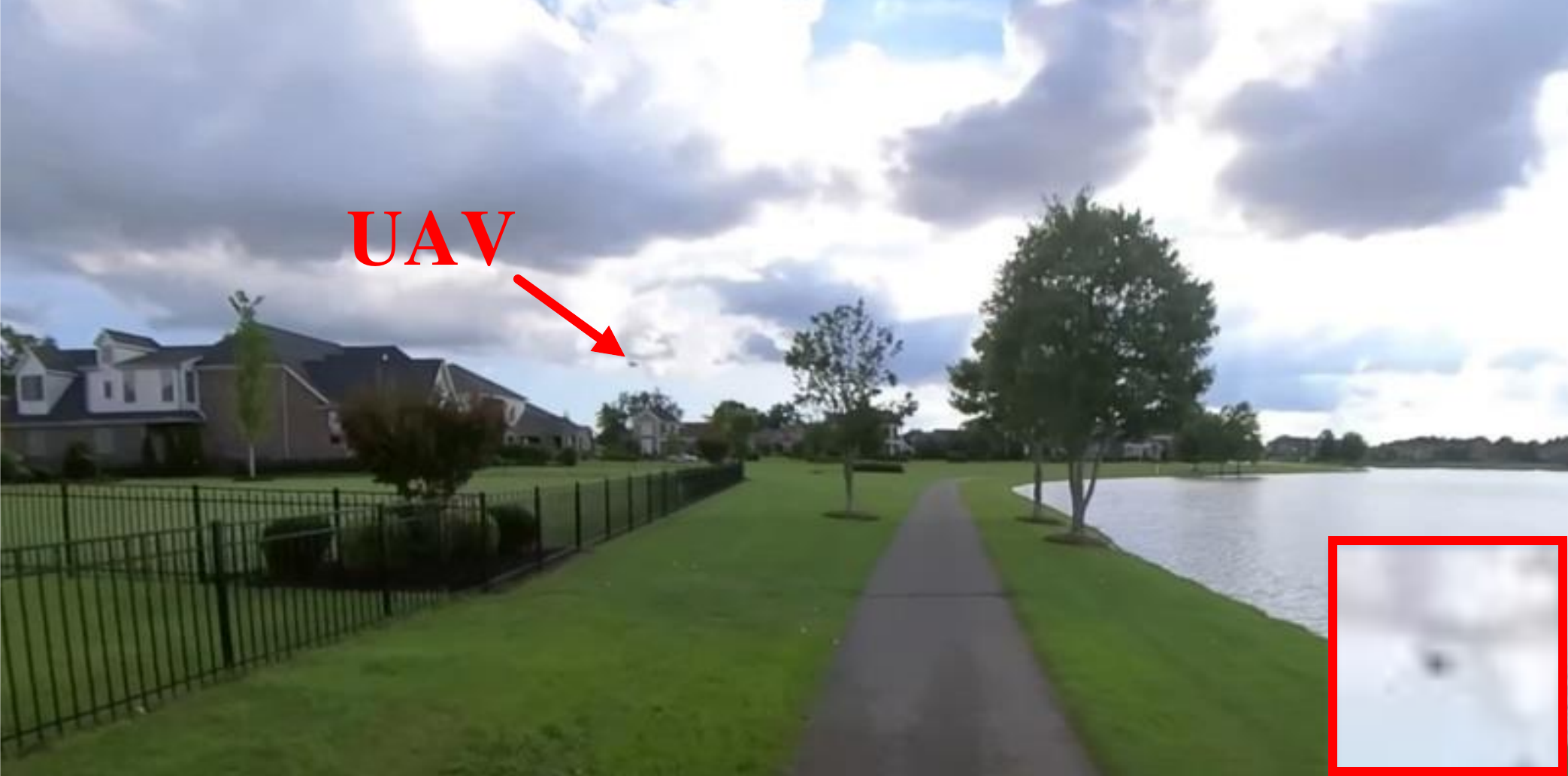}}
	\hfil
	\subfloat[]{\includegraphics[width=0.24\textwidth]{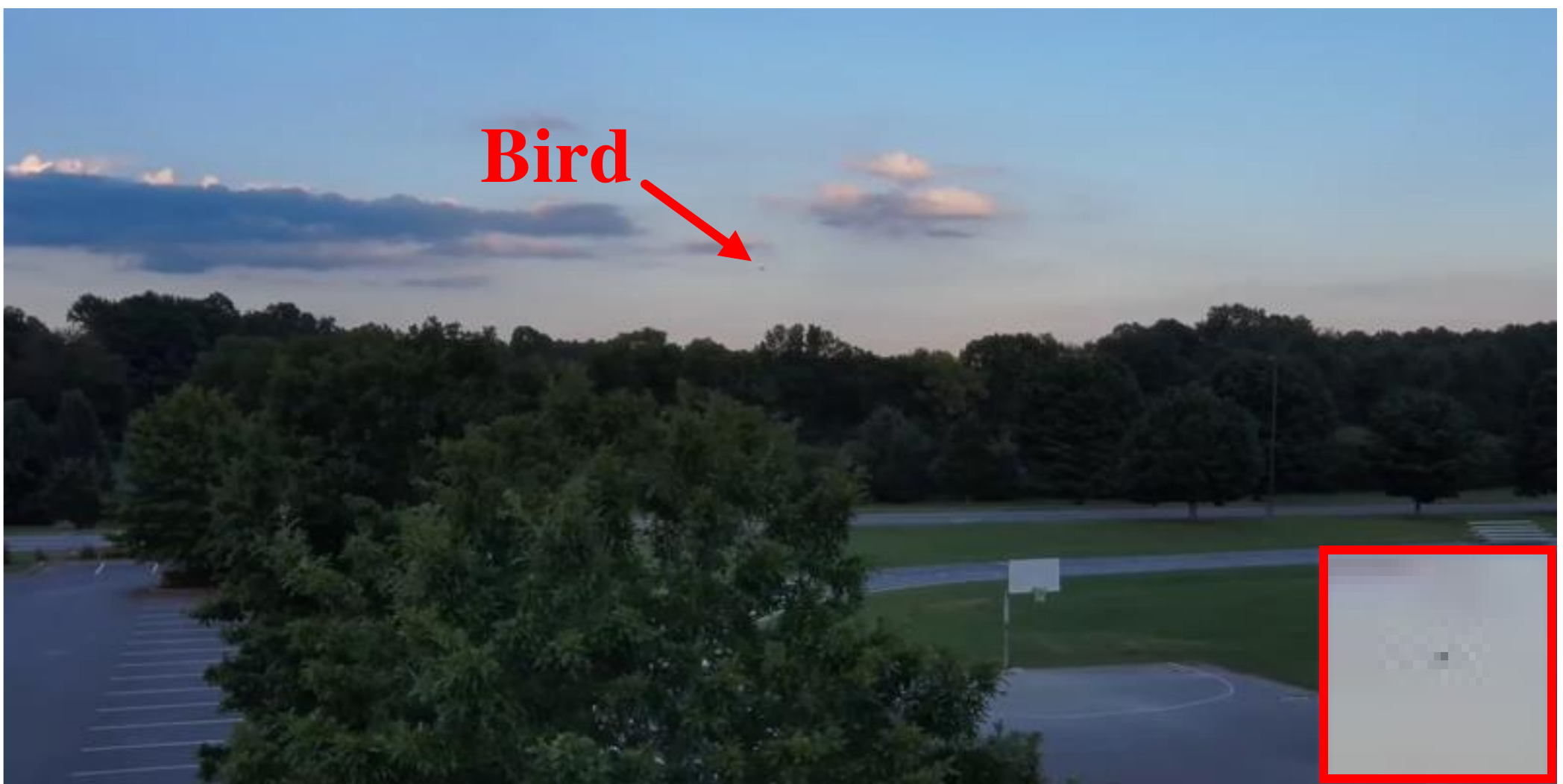}}
	\caption{Examples of small moving targets \cite{Youtube-UAV-Bird}. (a) An unmanned aerial vehicle (UAV), and (b) a bird in the distance. In each subplot, enlargements of the objects are shown in the red boxes. Both the UAV and bird appear as dim speckles, only a few pixels in size, and most of their visual features are difficult to discern. In particular, they all show extremely low contrast against the complex backgrounds.}
	\label{Examples-of-Small-Target}
\end{figure}

\IEEEPARstart{I}{n} the visual world, object motion provides important information to guide the behavior of observers (animals or robots). In the future, autonomous robotic systems will need to operate in complex dynamic environments, detecting object motions, understanding movement intention, predicting future paths, and reacting appropriately \cite{qiao2021survey, semnani2020force, yu2019bayesian}. It is accepted that detecting potentially dangerous objects early and far-off would permit sufficient time for responses to be made by autonomous systems, enabling them to maintain or enhance a dominant position in interaction and/or competition. However, if an object is extremely small or distant to the observer, it will always appear as a minute, dim speckle on the image, only one or a few pixels in size. Hence, most of the object's visual features will be difficult to determine, for example, an unmanned aerial vehicle (UAV) or a bird in the distance (Fig. \ref{Examples-of-Small-Target}).

Small target motion detection\footnote{Small target motion detection aims to detect moving object of interest that appear as a minute dim speckle on the image. Its size varies from one pixel to a few pixels, whereas other visual features, such as texture, orientation, and color, are difficult to recognize.} plays a critical role in a number of computer vision tasks including video surveillance, early warning, visual tracking and defence. For example, timely detection of  micro drones flying towards and over runways would help to protect airports from disruption. However, discriminating small moving targets in complex natural environments remains challenging to artificial visual systems. This is because: 1) small targets always equate to only a few pixels in size within images, presenting low-resolution appearance and unclear structure. Furthermore, most of the visual features, such as colour, orientation, and texture, are difficult to discern, which means feature representations of small targets necessary for motion detection are extremely weak; 2) small targets exhibit blurred boundaries and low contrast against heavily cluttered backgrounds, which makes them difficult to distinguish from noisy clutter; 3) freely moving camera could introduce complex changing scene and relative motion to small targets, which brings further challenges to motion discrimination.

Conventional approaches for motion detection can be classified into three main categories: frame difference \cite{saleemi2013multiframe}, background subtraction \cite{javed2018moving}, and optical flow \cite{fortun2015optical}. These approaches work best for static cameras, but their performance decreases significantly in applications involving mobile cameras, such as autonomous driving systems, flying drones, and mobile robots. In addition, these approaches cannot be directly applied to small target detection in complex natural environments, because: 1) they are unable to discriminate small targets from large objects in images, for example, pedestrian and/or vehicles; 2) small targets are always hidden within pixel errors and background noise after compensating for camera motion. Appearance-based methods can also be adopted for motion detection. Utilizing machine learning algorithms, such as convolutional neural networks \cite{Redmon_2016_CVPR}, support vector machines \cite{tang2017multiview} and evolutionary computation \cite{zhang2011evolutionary}, these methods classify moving objects based on the extracted low-level visual and high-level semantic features. However, they are ineffective against small objects that are only a few pixels in size, since most of visual features are hard to discern from low-resolution appearance of the objects.

Learning from the visual systems of animals provides a promising approach to build effective and robust models for detecting small moving targets in complex natural environments  \cite{wang2020bioinspired,sun2020decentralised,fu2019towards}. Despite the fact that the neural circuits in insects are relatively simpler than those in the human brains, insects achieve an extremely high success rate of $97\%$ in the pursuit of small flying mates or prey. The exquisite sensitivity of insects to small moving targets is supported by a class of specialized neurons called small target motion detectors (STMDs) \cite{nordstrom2006insect,barnett2007retinotopic,nordstrom2012neural}. These STMD neurons respond strongly to moving objects which occupy as small as $1^{\circ} - 3^{\circ}$ of the visual field, while exhibiting much weaker or even no response to large objects typically occupying more than $10^{\circ}$. In addition, the STMD neural responses are robust even when small targets display extremely low contrast against cluttered moving backgrounds. Understanding the biological neural computation that underlies small target motion detection would provide much needed inspirations for solving similar problems in autonomous systems.

Motivated by the superior properties of STMD neurons, several attempts have been made to develop quantitative STMD-based models for small target motion detection. Wiederman \emph{et al.} \cite{wiederman2008model} designed an Elementary STMD model (ESTMD) to detect the presence of small moving targets by multiplying luminance increase and decrease signals at each pixel after lateral inhibition. To determine motion direction of small targets, the Cascaded Model \cite{wiederman2013biologically} and Directionally Selective STMD (DSTMD) \cite{wang2018directionally} were developed by considering the correlation of luminance change signals from two different pixels. Wang \emph{et al.} \cite{wang2019Robust} proposed a visual system called STMD Plus, which takes into account both motion information and directional contrast, to filter out false positives in cluttered moving backgrounds. However, these models are heavily reliant on contrast between small targets and the background. As a result, their detection performance will degrade significantly as the target contrast decreases. In complex natural environments where small targets always exhibit extremely low contrast, it is difficult for these models to discern small target motion effectively and robustly.

To overcome these limitations, we develop an attention and prediction guided visual system (called apg-STMD). Prediction and attention are fundamental functions in the visual systems of insects, where the former utilizes present and/or past information to anticipate future object motion, while the latter prioritizes objects of interest amidst a swarm of potential alternatives \cite{nityananda2016attention,schroger2015bridging,bagheri2020evidence,wiederman2017predictive}. In the proposed visual system, an attention module and a prediction module are connected with an STMD-based neural network in a recurrent architecture. At each time step, the input image and a prediction map are applied to the attention module to search for potential small targets in several predicted areas. A contrast-enhanced image is produced by enhancing the contrast of potential targets over the input image, and then fed into the STMD-based neural network for discriminating small moving targets. The prediction module anticipates future positions of the detected small targets and generates a prediction map which is propagated to the attention module in the next time step. Experiments demonstrate the superior performance of the proposed visual system in detecting small target motion against complex backgrounds.

The remainder of this paper is organized as follows. Section \ref{Related-Work} discusses related research on motion-sensitive neural models, attention mechanism,  and prediction mechanism. We describe the proposed attention and prediction guided visual system in Section \ref{Formulation-of-the-System}. The experimental results on both synthetic and real-world data sets are reported in Section \ref{Results-and-Discussions}. Finally, Section \ref{Conclusion} concludes this paper.

\begin{figure*}[t!]
	\centering
	\includegraphics[width=0.98\textwidth]{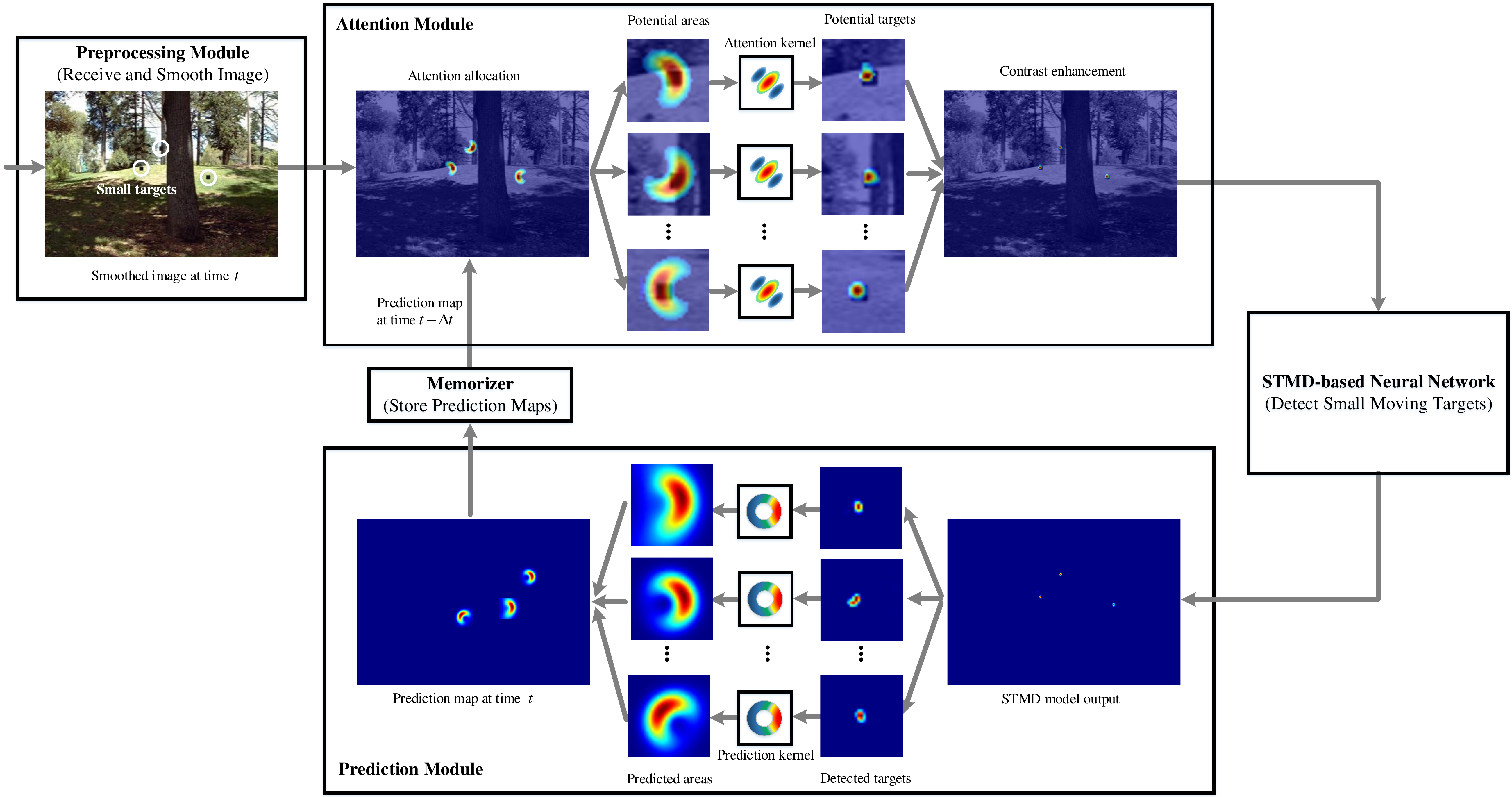}
	\caption{Overall flowchart of the proposed attention and prediction guided visual system. It consists of a preprocessing module (left), an attention module (top), an STMD-based neural network (right), a prediction module (bottom), and a memorizer (middle).}
	\label{Overall-Flowchart-of-Visual-System}
\end{figure*}

\section{Related Work}
\label{Related-Work}

\subsection{Motion-sensitive Neural Models}
The lobula giant movement detector (LGMD) \cite{rind2016two,rind1996neural}, lobula plate tangential cell (LPTC) \cite{maisak2013directional,perry2017generation}, and small target motion detector (STMD) \cite{nordstrom2006insect,barnett2007retinotopic,nordstrom2012neural} are three types of motion-sensitive neurons that have been widely investigated in the visual systems of insects. The LGMD responds most strongly to approaching objects, but shows little or no response to receding objects. It has been modelled as a collision detector that is further embodied in  micro mobile robots \cite{yue2006collision,yue2013redundant,hu2016bio,fu2019robust} and UAVs \cite{zhao2021enhancing,salt2019parameter} for collision avoidance. The LPTC is sensitive to objects which occupy a wide region of the visual field and which move in preferred directions. A wide-field LPTC can be modelled by an array of Hassenstein-Reichardt correlators \cite{eichner2011internal}, each of which focuses on a small part of the visual field. The LPTC model has been used for velocity estimation \cite{cope2016model}, collision avoidance \cite{bertrand2015bio}, and object tracking \cite{missler1995neural}. Although the LGMD and LPTC models perform well in detecting collision and wide-field motion, they are unable to discriminate small targets from other large objects in the visual field. 

The STMD gives peak responses to small moving targets that occupy only a few degrees of the visual filed, but much weaker responses to background movement and wide-field motion. The STMD-based models, such as ESTMD  \cite{wiederman2008model}, the Cascaded Model \cite{wiederman2013biologically}, DSTMD \cite{wang2018directionally}, and STMD Plus \cite{wang2019Robust}, have been developed to discriminate small moving targets against complex backgrounds. However, these models are all sensitive to contrast of small targets and perform poorly in complex natural environments where small target always exhibit extremely low contrast against their neighbouring backgrounds.

\subsection{Attention Mechanism}
Attention mechanism is fundamentally important for animals to forage, avoid predators, and search for mates. It focuses limited computation resources on parts of the visual field \cite{nityananda2016attention}. For example, bumblebees are able to select flowers of particular colours, while ignoring differently coloured distractors during visual searches \cite{nityananda2013bumblebee}; Drosophila selectively fixate on the most salient one in the swarms of prey and conspecifics that display different contrast against complex background \cite{sareen2011attracting}; fiddler crabs adjust their escape behaviour and selectively suppress neural responses to less dangerous predators when confronted with multiple threats in order to minimise the combined risk \cite{bagheri2020evidence}.

Attention mechanism has been commonly employed in computer modelling tasks such as image classification \cite{mnih2014recurrent}, visual question answering \cite{yang2016stacked}, natural language processing \cite{vaswani2017attention}, and image captioning \cite{chen2017sca}. It boosts model performance by adaptively choosing a sequence of regions for fine processing. However, it has not been utilized in artificial visual systems to detect small moving targets against complex natural backgrounds.  Moreover, the interaction of attention with prediction mechanisms for small target motion detection has not been investigated in depth.

\subsection{Prediction Mechanism}
Prediction mechanism plays a significant role in the visual systems of insects by anticipating future positions of prey and mates, and also contributing to path planning during rapid pursuit\cite{mischiati2015internal}. Recent research \cite{wiederman2017predictive} reveals that prediction process is able to enhance localized sensitivity to a small target ahead of its motion path, while exhibiting suppression elsewhere. Furthermore, when the target is occluded or abruptly disappears, the localized sensitivity will move forward and gradually weaken over time.  

The ability to model the prediction mechanisms of animals and use them to understand object motion in complex environment is extremely valuable for a wide range of applications. For example, reliably predicting the motions of surround objects (e.g., vehicles, pedestrians and cyclists) is a key requirement in the development of safe advanced autonomous driving technology  \cite{kooij2019context}; keeping track of current and future motion states of people is critical for socially-aware robots to avoid collision in populated environments \cite{luber2010people};  nursing-care assistant robots should be able to automatically anticipate human intentions by their actions to improve coordination and functionality \cite{stulp2015facilitating}. However, little work has been done on modelling prediction mechanisms to anticipate small target motion against complex natural backgrounds.

\section{Attention and Prediction Guided Visual System}
\label{Formulation-of-the-System}
The proposed visual system is composed of five subsystems, including three modules (preprocessing, attention, prediction), an STMD-based neural network, and a memorizer, as illustrated in Fig. \ref{Overall-Flowchart-of-Visual-System}. Once an image is received at time $t$, it is first smoothed by the preprocessing module, then applied to the attention module to determine several potential areas based on the prediction map from the memorizer. In each area, potential small targets are selected by convolution with the attention kernels and their contrast to background is enhanced by addition of the convolutional outputs. The contrast-enhanced image is fed into the STMD-based neural network for discriminating small moving targets from complex background. In the prediction module, futures positions of the detected targets at time $t+\Delta t$ are anticipated by convolution with the prediction kernels, then merged into a prediction map that is stored in the memorizer for next input image. We introduce network architecture of the proposed visual system in Section \ref{Architecture of the Visual System}, then describe its components in Section \ref{Preprocessing Module} -- \ref{Prediction Module}. 

\subsection{Network Architecture of the Proposed Visual System}
\label{Architecture of the Visual System}

\begin{figure}[!t]
	\centering
	\includegraphics[width=0.45\textwidth]{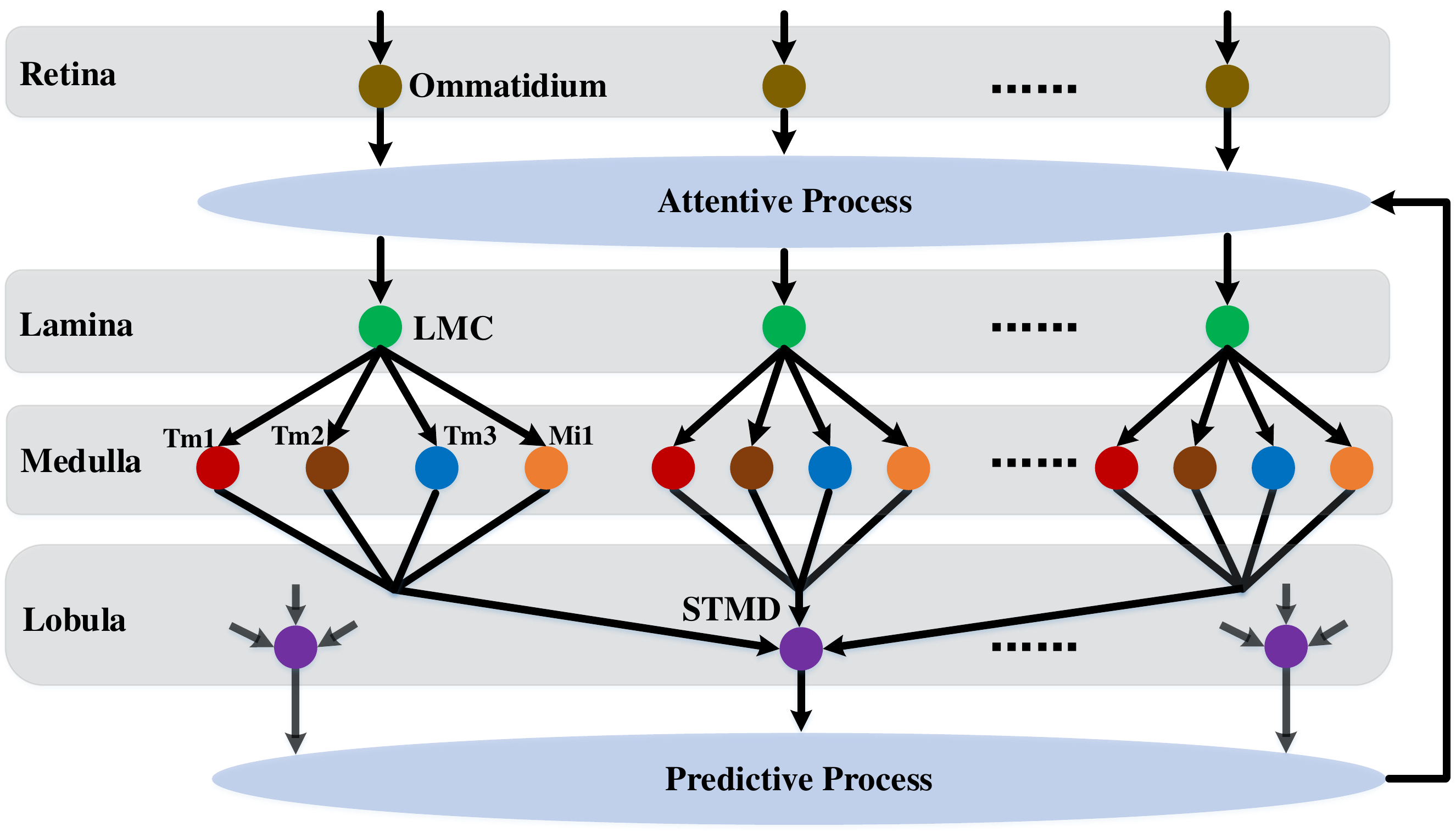}
	\caption{Network architecture of the proposed attention and prediction guided visual system. Each neuron is represented by a coloured circular node.}
	\label{Network-architecture-of-Visual-System}
\end{figure}

To realize the functions in Fig. \ref{Overall-Flowchart-of-Visual-System}, a number of specialized neurons are coordinated in the proposed visual system whose network architecture is shown in Fig. \ref{Network-architecture-of-Visual-System}. As can be seen, the proposed visual system is composed of four neural layers, including retina, lamina, medulla, and lobula \cite{behnia2014processing}, where an attention and a prediction mechanisms are implemented on the outputs of the retina and lobula, respectively. Specifically, ommatidia \cite{warrant2017remarkable} capture and preprocess visual information from the whole scene, then the attention is allocated to parts of the visual scene to enhance signals of potential small targets. The enhanced signals are applied to large monopolar cells (LMCs) \cite{freifeld2013gabaergic}, further parallelly processed by four medulla neurons (i.e., Tm1, Tm2, Tm3, and Mi1) \cite{takemura2013visual}, finally integrated in the STMDs to detect small target motion within complex natural environments. Future positions of the detected targets are predicted and then fed back to the attentive process.

\subsection{Preprocessing Module}
\label{Preprocessing Module}
The functionalities of the preprocessing module is implemented by numerous ommatidia located in the retina layer, as depicted in Fig. \ref{Network-architecture-of-Visual-System}. To receive an entire image as model input, the preprocessing module first arranges ommatidia in matrix form. Then the luminance of each pixel is captured by each ommatidium whose sensitivity function is modelled as a Gaussian kernel \cite{caves2018visual}. Formally, we represent input image by $I(x,y,t) \in \mathbb{R}$ where $(x,y)$ is spatial coordinates while $t$ denotes time. Given a Gaussian kernel with standard deviation $\sigma_1$
\begin{equation}
	G_{\sigma_1}(x,y)= \frac{1}{2\pi\sigma_1^2}\exp(-\frac{x^2+y^2}{2\sigma_1^2})
	\label{Photoreceptors-Gauss-blur-Kernel}
\end{equation}
then the output of an ommatidium $P(x,y,t)$ is defined as
\begin{equation}
	P(x,y,t) =  \iint I(u,v,t)G_{\sigma_1}(x-u,y-v)dudv.
	\label{Photoreceptors-Gaussian-Blur}
\end{equation}

\subsection{Attention Module}
\label{Section-Attention-Module}

\begin{figure}[t!]
	\centering
	\subfloat[]{\includegraphics[width=0.25\textwidth]{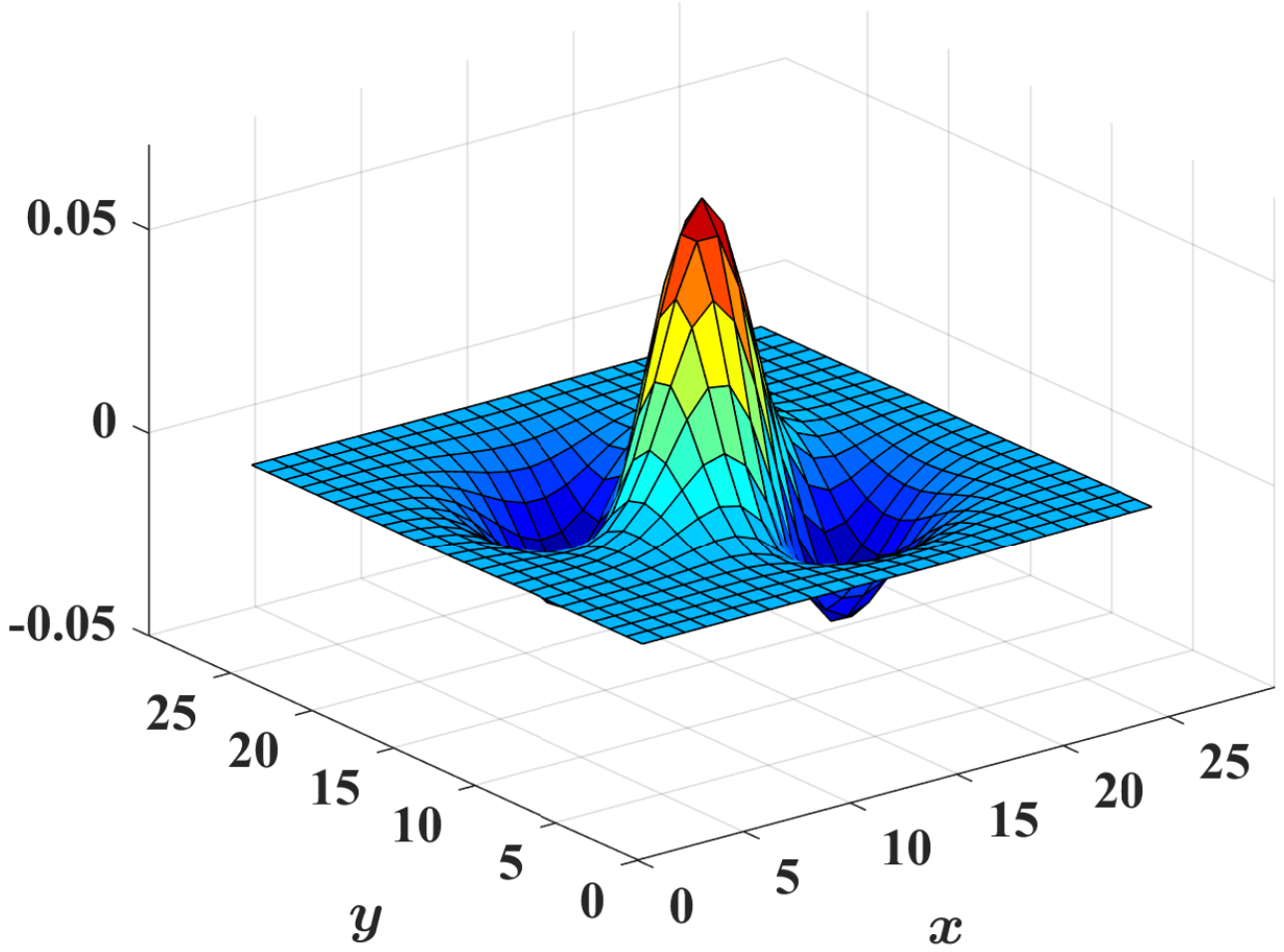}}
	\hfil
	\subfloat[]{\includegraphics[width=0.20\textwidth]{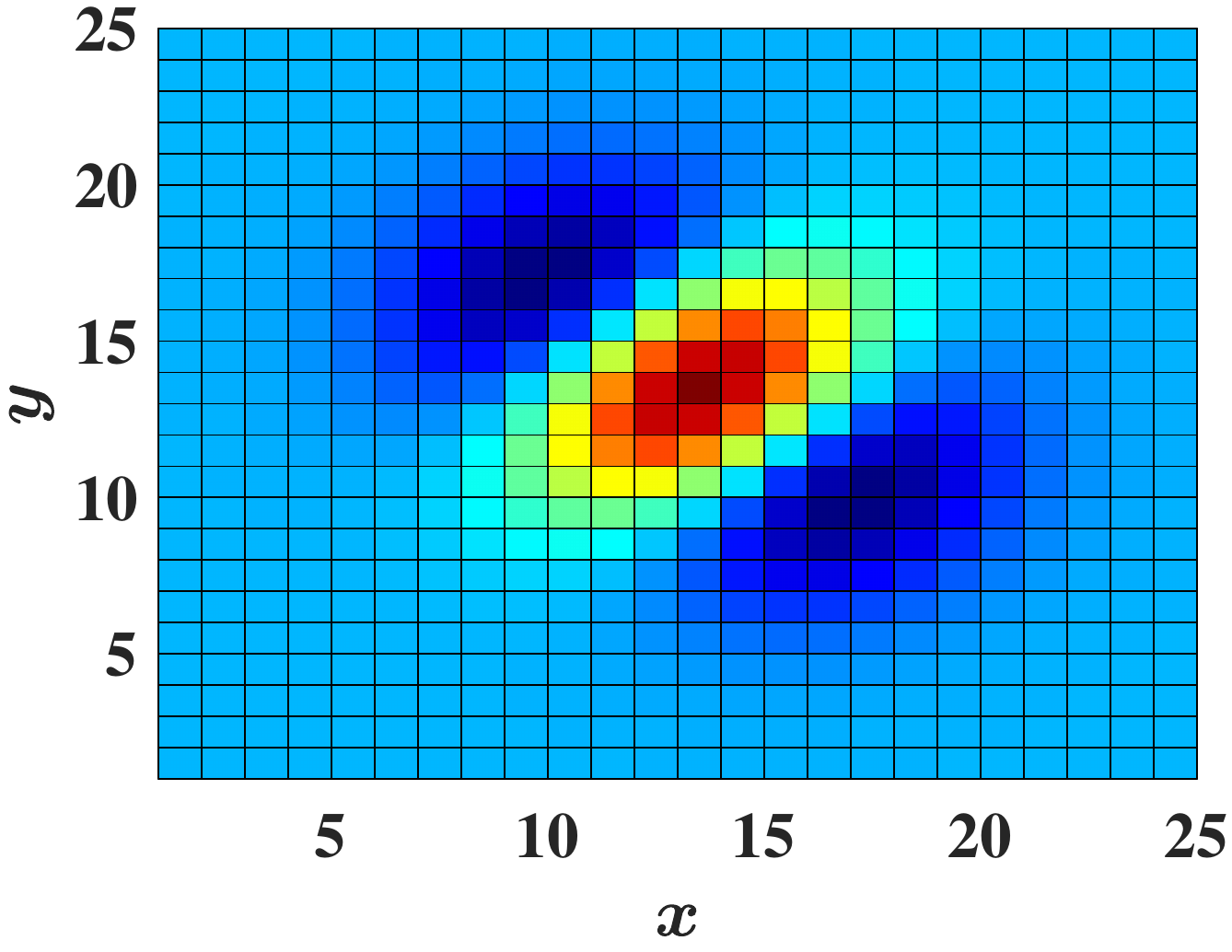}}
	\caption{(a) Three-dimensional and (b) planar representations of an attention kernel $W_a(x,y,\varsigma,\theta)$ where $\varsigma= 3.0$, $\theta = \pi/4$.}
	\label{Illustration-of-Attention-Kernel}
\end{figure}

As can be seen from Fig. \ref{Overall-Flowchart-of-Visual-System}, a smoothed image $P(x,y,t)$ and a prediction map $M(x,y,t-\Delta t)$ form the inputs of the attention module. The prediction map is initialized to zero and updated recursively in the prediction module. A set of potential areas denoted as $\{\Omega_i | i=1,2,\cdots,N\}$ is first determined by comparing $M(x,y,t-\Delta t)$ with a preset threshold. To search for potential small targets in each area $\Omega_i$, the attention module convolves $\Omega_i$ with a family of attention kernels. Let $\Sigma$ and $\Theta$ denote sets of scale and orientation, respectively, then an attention kernel is defined as
\begin{equation}
	W_a(x,y,\varsigma,\theta)= 2\frac{\varsigma^2-(x\cos\theta + y\sin\theta)^2}{\pi\varsigma^4}\exp(-\frac{x^2+y^2}{2\varsigma^2})
	\label{Attention-Kernel}
\end{equation}
where scale $\varsigma \in \Sigma$ and orientation $\theta \in \Theta$. As shown in Fig. \ref{Illustration-of-Attention-Kernel}, the attention kernel measures the luminance difference between the central part and surrounding areas on both sides along the orientation $\theta$. Since a small target always displays speckle-like structure in an image whose luminance is higher or lower than that of its surrounding background, a significant response will appear at the target position after the convolution with an attention kernel. To suppress non-speckle structures, such as lines, edges, and corners, we select the minimal convolution output by varying kernel orientation $\theta$ for each scale $\varsigma$, then obtain the maximal output among all the scales \cite{Wang_2017_ICCV}, that is
\begin{equation}
	A_i(x,y,t) =  \max_{\varsigma \in \Sigma} \min_{\theta \in \Theta} \iint_{\Omega_i} P(u,v,t)W_a(x-u,y-v,\varsigma,\theta)dudv
	\label{}
\end{equation}
where $A_i(x,y,t)$ denotes the response of the attention module in the local area $\Omega_i$. To enhance the contrast of  potential targets against their surrounding backgrounds, we add $A_i(x,y,t)$ with the smoothed image $P(x,y,t)$, that is
\begin{equation}
	P_e(x,y,t) =  P(x,y,t) + \alpha \sum_{i=1}^N A_i(x,y,t)
\end{equation}
where $P_e(x,y,t)$ denotes the contrast-enhanced image, $\alpha$ is a constant, and $N$ is the number of the local areas $\Omega_i$.

\subsection{STMD-based Neural Network}

\begin{figure}[!t]
	\centering
	\includegraphics[width=0.48\textwidth]{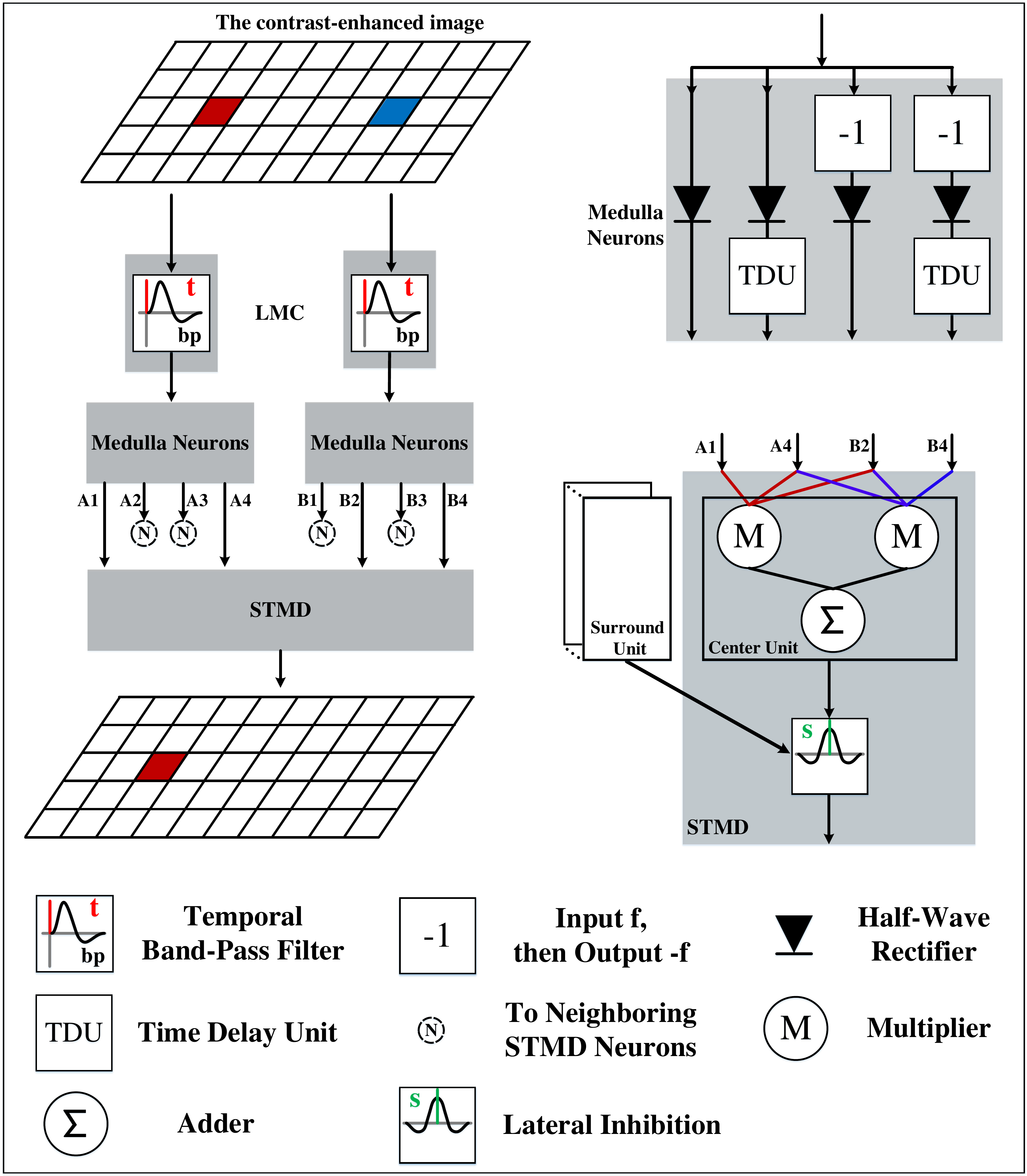}
	\caption{Schematic of the STMD-based neural network. At each time step, it receives an entire contrast-enhanced image from the attention module as the network input which is processed by the LMCs, medulla neurons, and STMDs sequentially. Note that only a STMD and its pre-synaptic neurons are presented here for clarity, but they are all arranged in matrix form.}
	\label{Schematic-of-the-STMD-based-Neural-Network}
\end{figure}

The STMD-based neural network consists of three sequentially arranged neural layers, including lamina, medulla, and lobula, as shown in Fig. \ref{Network-architecture-of-Visual-System}. To detect small moving targets against complex natural background, the contrast-enhanced image $P_e(x,y,t)$ from the attention module is processed by the LMCs, medulla neurons, and STMDs in a feedforward manner.

\subsubsection{Large Monopolar Cells}
Luminance of a pixel will change over time when an object passes through it. To measure temporal changes in luminance of each pixel, we model the LMC as a band-pass filter in the time domain (Fig. \ref{Schematic-of-the-STMD-based-Neural-Network}). Considering excellent temporally-processing features of Gamma kernel \cite{de1991theory}, we adopt the difference of two Gamma kernels as the impulse response of the temporal filter $H(t)$
\begin{align}
H(t) &= \Gamma_{n_1,\tau_1}(t) - \Gamma_{n_2,\tau_2}(t) \label{BPF-Para}\\
\Gamma_{n,\tau}(t) &= (nt)^n \frac{\exp(-nt/\tau)}{(n-1)!\cdot \tau^{n+1}}
\end{align}
where $\Gamma_{n,\tau}(t)$ represents a Gamma kernel and its temporal response characteristics are completely determined by order $n$ and time constant $\tau$. The output of a LMC is given by convolution of $H(t)$ with the contrast-enhanced image $P_e(x,y,t)$
\begin{equation}
L(x,y,t) = \int P_e(x,y,s)H(t-s) ds
\label{LMCs-HPF}
\end{equation}
where $L(x,y,t)$ denotes the output of a LMC corresponding to pixel $(x,y)$ at time $t$. Note that $L(x,y,t)$  discloses the changes in luminance at pixel $(x,y)$ with respect to time $t$. Specifically, a positive output means an increase in luminance whereas a negative one reflects a decrease in luminance.

\subsubsection{Medulla Neurons}
Four medulla neurons, including Tm1, Tm2, Tm3, and Mi1, are connected to a single LMC and process the output of the LMC $L(x,y,t)$ in parallel, as can be seen from Fig. \ref{Network-architecture-of-Visual-System}. More precisely, the Tm3 serves as a half-wave rectifier to allow the positive part of $L(x,y,t)$ while blocking the negative part; in contrast, the Tm2 allows the negative part and blocks the positive part. Let $S^{\text{Tm3}}(x,y,t)$ and $S^{\text{Tm2}}(x,y,t)$ denote the output of the Tm3 and Tm2, respectively, then they can be formulated as
\begin{align}
	S^{\text{Tm3}}(x,y,t) &= [L(x,y,t)]^{+}  \label{Tm3-Output} \\
	S^{\text{Tm2}}(x,y,t) &= [-L(x,y,t)]^{+} \label{Tm2-Output}
\end{align}
where $[x]^+$ refers to $\max (x,0)$. As shown in Fig. \ref{Schematic-of-the-STMD-based-Neural-Network}, the Mi1 and Tm1 neurons serve as half-wave rectifiers followed with a time-delay unit (TDU) where the temporal delay is implemented by convolution with a Gamma kernel. Denote the outputs of the Mi1 and Tm1 as $S_{{(n,\tau)}}^{\text{Mi1}}(x,y,t)$ and $S_{{(n,\tau)}}^{\text{Tm1}}(x,y,t)$, then they can be described as
\begin{align}
	S_{{(n,\tau)}}^{\text{Mi1}}(x,y,t) &= \int [L(x,y,s)]^{+} \cdot \Gamma_{n,\tau}(t-s) ds \label{Mi1-Output}\\
	S_{{(n,\tau)}}^{\text{Tm1}}(x,y,t) &= \int [-L(x,y,s)]^{+} \cdot  \Gamma_{n,\tau}(t-s) ds \label{Tm1-Output}
\end{align}
where order $n$ and time constant $\tau$ of Gamma kernel $\Gamma_{n,\tau}(t)$ control the time-delay order and length, respectively.

\subsubsection{Small Target Motion Detectors}
\begin{figure}[!t]
	\centering
	\includegraphics[width=0.18\textwidth]{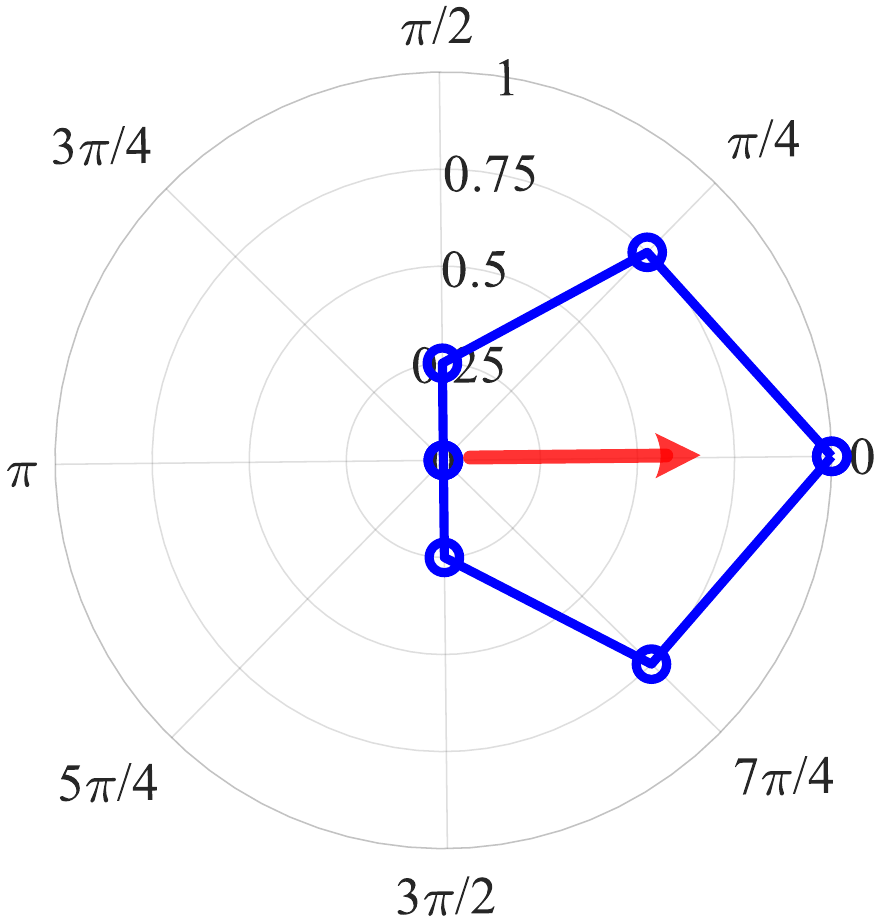}
	\caption{Normalized outputs of the STMD neuron to a small target at pixel $(x_0,y_0)$ and time $t_0$ along eight preferred directions $\theta \in \{0,\frac{\pi}{4},\frac{\pi}{2},\frac{3\pi}{4},\pi, \frac{5\pi}{4},\frac{3\pi}{2},\frac{7\pi}{4}\}$. The red arrow represents the object's motion direction.}
	\label{DSTMD-Model-Outputs-Polar}
\end{figure}

\begin{figure}[!t]
	\centering
	\includegraphics[width=0.30\textwidth]{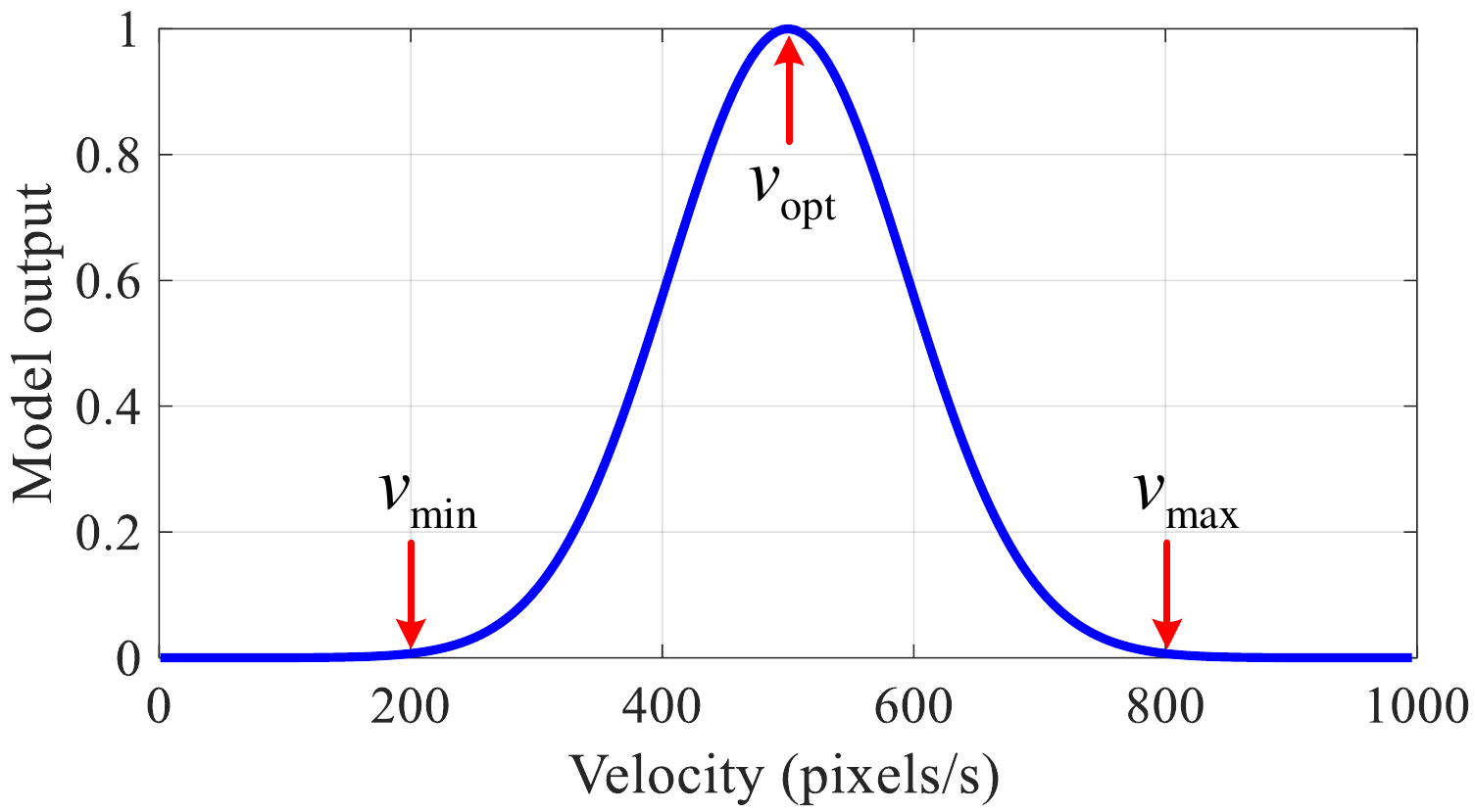}
	\caption{Normalized outputs of the STMD neuron to a small target with different velocities. $[v_{\text{min}},v_{\text{max}}]$ and $v_{\text{min}}$ denote the preferred velocity range and the optimal velocity of the STMD, respectively.}
	\label{Illustration-of-Velocity-Selectivity}
\end{figure}

Medulla neurons at two different pixels provide inputs to an STMD neuron, as illustrated in Fig. \ref{Schematic-of-the-STMD-based-Neural-Network}. The two pixels denoted by $(x,y)$ and  $(x'(\theta),y'(\theta))$, respectively, are formulated as
\begin{equation}
	\begin{split}
		x'(\theta) &= x + \gamma \cdot \cos\theta \\ 
		y'(\theta) &= y + \gamma \cdot \sin\theta
	\end{split}
	\label{DSTMD-Signal-Correlation-Distance}
\end{equation}
where $\theta$ represents the preferred direction of the STMD, $\gamma$ denotes a constant. Note that when an object moves from pixel $(x,y)$ to $(x'(\theta),y'(\theta))$, it will induce increase and decrease in luminance of the two pixels. The luminance-change information have been captured by the LMCs, and further separated into increasing and decreasing components by the four medulla neurons. To produce a significant response to the moving object, the STMD first aligns these luminance-increase and luminance-decrease signals correctly in the time domain, and then multiplies the temporally-aligned signals together \cite{wang2018directionally}, that is
\begin{equation}
	\begin{split}
		D(x, y,t,\theta) &= S^{\text{Tm3}}(x,y,t) \times \\  
		& \Big \{S^{\text{Tm1}}_{{(n_{_4},\tau_{_4})}}(x,y,t) +S^{\text{Mi1}}_{{(n_{_3},\tau_{_3})}}(x'(\theta),y'(\theta),t)\Big\} \\
		& \times S^{\text{Tm1}}_{{(n_{_5},\tau_{_5})}}(x'(\theta),y'(\theta),t)
		\label{DSTMD-Signal-Correlation}
	\end{split}
\end{equation}
where $D(x,y,t,\theta)$ represents the output of the STMD with a preferred direction $\theta$; the time constants, i.e., $\tau_3$, $\tau_4$, and $\tau_5$, are determined by time intervals between the luminance-increase and decrease signals of the two pixels; the orders, i.e., $n_3$, $n_4$, and $n_5$, control the shapes of signals after temporal alignment.

The correlation output $D(x,y,t,\theta)$ is further convolved with two inhibition kernels, including $W_s(x,y)$ in the spatial domain for suppressing responses to large moving objects, and $W_d(\theta)$ in the direction domain for inhibiting responses of more than $45^{\circ}$ apart, which are defined as
\begin{align}
	W_s(x,y)  &= A  \cdot [g(x,y)]^{+}  + B \cdot [g(x,y)]^{-} \label{Lateral-Inhibition-Kernel-1} \\
	g(x,y)  & = G_{\sigma_2}(x,y) - e \cdot G_{\sigma_3}(x,y) - \rho \label{Lateral-Inhibition-Kernel-2} \\
	W_d(\theta) & = G_{\sigma_4}(\theta) - G_{\sigma_5}(\theta)       \label{Directional-Inhibition-Kernel}
\end{align}
where $[x]^+$ and $[x]^-$ refer to $\max (x,0)$ and $\min (x,0)$, respectively; $A$, $B$, $e$, and $\rho$ are constant. The output of the STMD after the inhibition $E(x,y,t,\theta)$ is described as
\begin{equation}
\begin{split}
E(x,y,t,\theta) = \iiint &D(u,v,t,\psi) \cdot W_s(x-u,y-v) \\
 & \cdot W_d(\theta - \psi) du dv d\psi.
\end{split}
\label{DSTMD-Lateral-Inhibition}
\end{equation}
Fig. \ref{DSTMD-Model-Outputs-Polar} shows $E(x,y,t,\theta)$ at pixel $(x_0,y_0)$ and time $t_0$ along eight preferred directions $\theta$.  As can be seen, $E(x,y,t,\theta)$ is directionally selective. Specifically, the strongest response appears at the motion direction of the small target, i.e.,  $\theta = 0$. When the preferred direction deviates from $\theta = 0$, the neural output decreases significantly and equates to zero at $\theta = \pi$ opposite to the motion direction. In addition, $E(x,y,t,\theta)$ exhibits strong velocity selectivity, as illustrated in Fig. \ref{Illustration-of-Velocity-Selectivity}. Specifically, the STMD responds to small targets with velocities in a specific range denoted by $[v_{\text{min}},v_{\text{max}}]$, and its output peaks at an optimal velocity denoted by $v_{\text{opt}}$. Note that direction and velocity selectivities have been found in real STMD neurons \cite{nordstrom2006insect}.

To determine locations and motion directions of small targets, we compare $E(x,y,t,\theta)$ with a threshold $\delta$. Specially, if $E(x,y,t,\theta) > \delta$, then we consider $(x,y,t,\theta)$ as a positive detection which means a small target that moves along direction $\theta$ is detected at pixel $(x,y)$ and time $t$. However, $E(x,y,t,\theta)$ may contain a number of false positives induced by small-target-like features in complex backgrounds. To eliminate these false positives, we adopt the method proposed in \cite{wang2019Robust}. Specifically, true positives are distinguished from false positives by comparing variation amount of directional contrast on their motion traces (represented by standard deviation). If directional contrast on the motion trace of a detected object varies significantly with time, then we believe that the detected object is a true positive; otherwise, it is a false positive.

\subsection{Prediction Module}
\label{Prediction Module}

\begin{figure}[t!]
	\centering
	\includegraphics[width=0.30\textwidth]{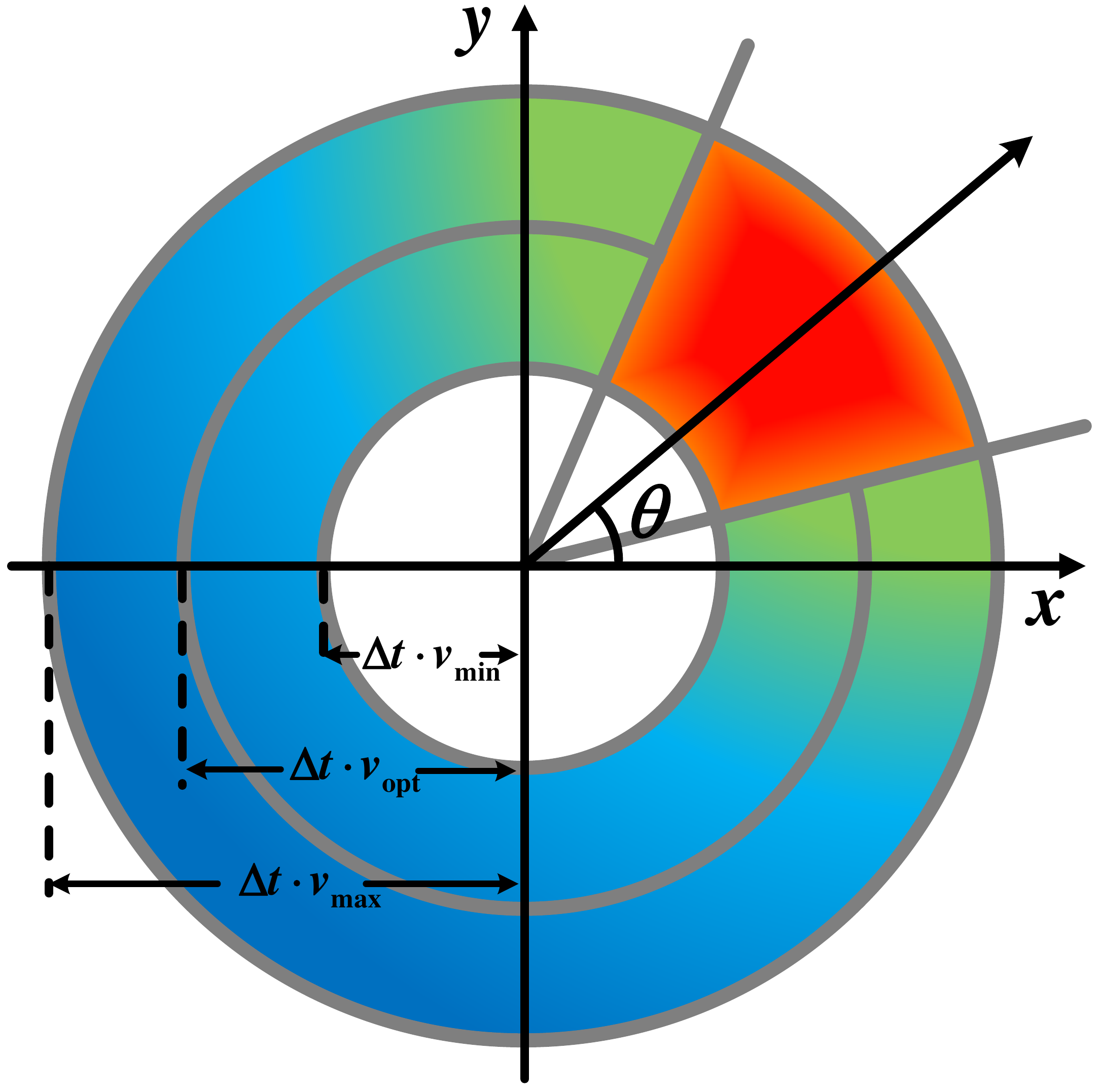}
	\caption{Potential positions of a small target at time $t+\Delta t$ where the origin coordinates are $(x_t,y_t)$ and $\theta$ denotes the motion direction. The red-ring area represents the positions where the small target would appear with a high probability.}
	\label{Motion-Prediction-Analysis}
\end{figure}

\begin{figure}[t!]
	\centering
	\subfloat[]{\includegraphics[width=0.25\textwidth]{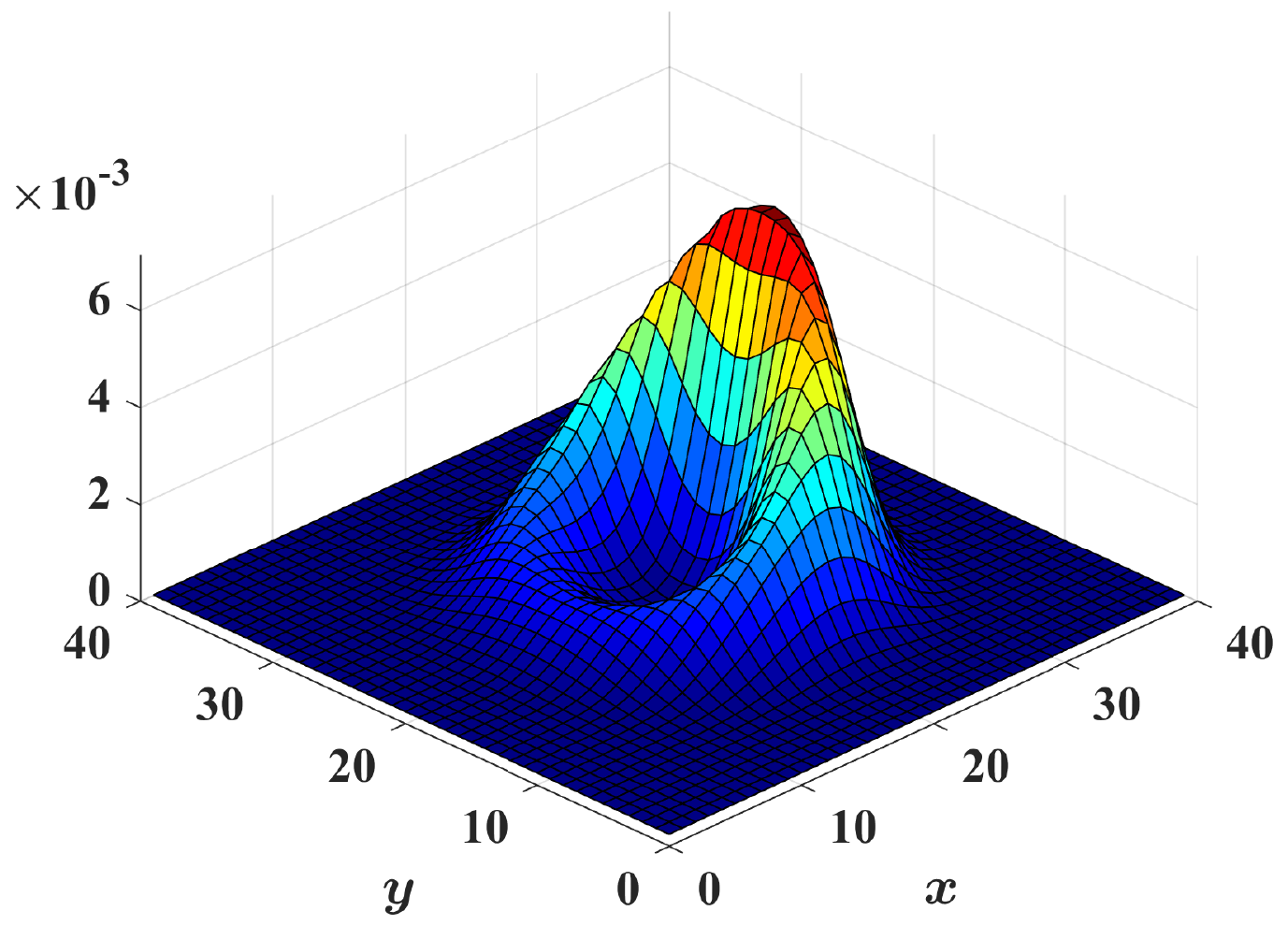}}
	\hfil
	\subfloat[]{\includegraphics[width=0.20\textwidth]{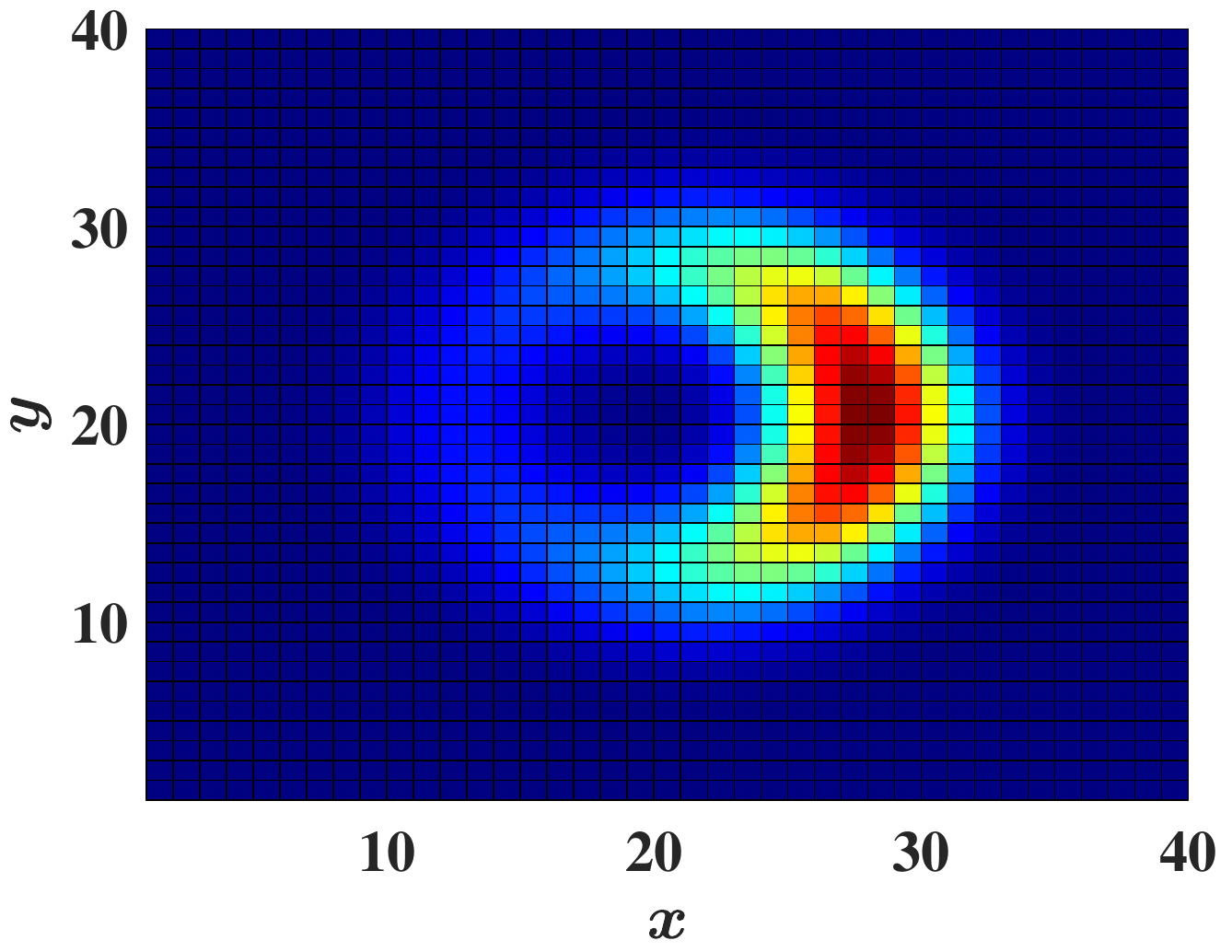}}
	\caption{(a) Three-dimensional and (b) planar representations of a prediction kernel $W_p(x,y,\theta)$ where $\theta = 0$.}
	\label{Prediction-Kernel-3D-2D}
\end{figure}

The spatial coordinates $(x_t,y_t)$ and motion directions $\theta_t$ of the small targets at time $t$ are obtained by the STMD-based neural network, then fed into the prediction module to anticipate their future positions. Let $(x_{t+ \Delta t},y_{t+ \Delta t})$ denote the position of a small target at time $t+\Delta t$, then it can be described as 
\begin{equation}
	\begin{split}
	(x_{t+ \Delta t},y_{t+ \Delta t}) =  & (x_t,y_t) \\
	 & + v_t(\cos(\theta_t + \omega_t),\sin(\theta_t + \omega_t))\Delta t   \\ 
	        & + \varepsilon_t(\cos(\theta_t + \omega_t),\sin(\theta_t + \omega_t)) \Delta t 
	\end{split}
\label{Formulation-of-Future-Positions}
\end{equation}
where $v_t$ denotes the velocity of the small target at time $t$; $\varepsilon_t$ and $\omega_t$ represent deviations of the velocity and motion direction over the time period $\Delta t$, respectively. Note that the STMD is selective to object velocity (see Fig. \ref{Illustration-of-Velocity-Selectivity}),  so $v_t$ is in the preferred velocity range of the STMD, i.e,  $v_{\text{min}} \leq v_t \leq v_{\text{max}}$. Moreover, if the velocity and motion direction of the small target have not significant changes over the period $\Delta t$, i.e., $\varepsilon_t$ and $\omega_t$ are all equal to low values in (\ref{Formulation-of-Future-Positions}), the small target would appear in a ring area with a high probability at time $t+\Delta t$, as depicted in Fig. \ref{Motion-Prediction-Analysis}. 

Based on the above observation, we define a set of prediction kernels $W_p(x,y,\theta)$ with various orientations $\theta$ as	
\begin{equation}
\begin{split}
W_p(x,y,\theta) = \lambda & \cdot \exp\Big(-\frac{(x-v_{\text{opt}} \cos\varphi \Delta t)^2}{2\zeta^2} \Big) \\
	&\cdot \exp\Big(-\frac{(y-v_{\text{opt}} \sin\varphi \Delta t)^2}{2\zeta^2}\Big) \\
& \cdot \exp\Big(\eta \cos(\varphi - \theta) \Big) 
\end{split}
\label{Motion-Prediction-Kernel}
\end{equation}
where $v_{\text{opt}}$ stands for the optimal velocity of the STMD; $\varphi$ denotes the angle between the vector $(x,y)$ and the positive direction of $x$-axis, $0\leq \varphi < 2\pi$; $\lambda$ represents normalization factor; $\zeta$ and $\eta$ are constant. As can be seen from Fig. \ref{Prediction-Kernel-3D-2D}, the shape of $W_p(x,y,\theta)$  is similar to that of the future positions in Fig. \ref{Motion-Prediction-Analysis}, which displays as a ring structure. In addition, the value of $W_p(x,y,\theta)$ reveals the probability of a small target appearing at pixel $(x,y)$. We further define the predictive gain of the STMD $F(x,y,t,\theta)$ by   
\begin{equation}
\begin{split}
F(x,y,t,\theta) = \iint \Big\{\mu &E(u,v,t,\theta) + (1-\mu)F(u,v,t-\Delta t,\theta) \Big\} \\
&\cdot W_p(x-u,y-v,\theta) dudv
\end{split}
\label{Motion-Prediction-Conv}
\end{equation}
where $E(x,y,t,\theta)$ is the output of the STMD at time $t$; $F(x,y,t-\Delta t,\theta)$ denotes the predictive gain at time $t-\Delta t$; $\mu$ is constant and $ 0 \leq \mu \leq 1$. To generate a prediction map, we integrate the predictive gain $F(x,y,t,\theta)$ in the direction domain, that is
\begin{equation}
   M(x,y,t)= \int F(x,y,t,\theta) d\theta
\end{equation}
where $M(x,y,t)$ denotes the prediction map at time $t$ which anticipates the locations of small targets at $t+\Delta t$. The predictive process is able to facilitate responses of the STMD \cite{wiederman2017predictive}, so we define the facilitated STMD output $Q(x,y,t,\theta)$  by summing the STMD neural output with the previous prediction gains, that is
\begin{equation}
\begin{split}
Q(x,y,t,&\theta) =   E(x,y,t,\theta) \\
               & + \beta \cdot \int_{t-\Delta t}^{t} e^{\kappa \cdot (t-s-\Delta t)} \cdot F(x,y,s,\theta) ds
\end{split}
\label{Motion-Predictive-Gain}
\end{equation}
where $\beta$, $\kappa$ are constant.

\subsection{Memorizer}

As shown in Fig. \ref{Overall-Flowchart-of-Visual-System}, the memorizer collects prediction maps $M(x,y,t)$ from the prediction module. Let $\{M(x,y,t)| t \in [0,t_c]\}$ denote the set of the prediction maps where $t_c$ stands for the current time step. For a new input image at time $t_c + \Delta t $, the memorizer provide the prediction map $M(x,y,t_c)$ to the attention module for determining potential areas.


\section{Experiments and Discussions}
\label{Results-and-Discussions}

\subsection{Experimental Setup}

\subsubsection{Data Sets}

We used a simulated data set (Vision Egg) \cite{straw2008vision} and a real-world data set (RIST) \cite{RIST-Data-Set} to evaluate the proposed model (apg-STMD) on small target motion detection task. The Vision Egg data set covers a wide variety of synthetic small targets exhibiting a range of luminance, velocity, and size, moving against complex backgrounds. Each synthetic video contains one or multiple small target motions, whose resolution and sampling frequency equate to $500 \times 250$ pixels and $1000$ Hz, respectively. The RIST data set contains $19$ videos captured in the wild using an action camera (GoPro Hero $6$) with a resolution $480 \times 270$ pixels at $240$ fps. High sampling rate is set to ensure that captured images are blur-free and every critical moment of object motion is recorded. The scenarios of recorded videos covers various challenges, such as highly complex dynamic backgrounds, low-contrast targets, illumination variations, bad weather conditions, and sudden background movements. Each video holds a small moving target whose size ranges between $3\times 3$ and $15 \times 15$ pixels.

\begin{table}[t!]
	\renewcommand{\arraystretch}{1.3}
	\caption{Parameters of the Proposed pag-STMD Model.}
	\label{Table-Parameter-pag-STMD}
	\centering
	\begin{tabular}{cc}
		\hline
		Eq. & Parameters \\	
		\hline
		(\ref{Photoreceptors-Gaussian-Blur}) & $\sigma_1 = 1$ \\
		
		(\ref{Attention-Kernel})           & $\Sigma = \{2.0,2.5,3.0,3.5\}$, $\Theta = \{0, \pi/4, \pi/2, 3\pi/4\}$ \\
		
		(\ref{BPF-Para}) & $n_1 = 2, \tau_1= 3, n_2 = 6,\tau_2 = 9$\\
		
		(\ref{DSTMD-Signal-Correlation-Distance}) & $\gamma = 3$ \\
		
		(\ref{DSTMD-Signal-Correlation}) & $n_3 = 3, \tau_3 = 15, n_4 = 5, \tau_4 = 25, n_5 = 8, \tau_5 = 40$ \\
		
		(\ref{Lateral-Inhibition-Kernel-1}) & $A = 1, B = 3.5$ \\
		
		(\ref{Lateral-Inhibition-Kernel-2}) & $\sigma_2 = 1.25, \sigma_3 = 2.5, e = 1.2, \rho = 0$ \\
		
		(\ref{Directional-Inhibition-Kernel}) & $\sigma_4 = 1.5, \sigma_5 = 3$ \\
		
		(\ref{Motion-Prediction-Kernel}) & $\zeta = 2, \eta = 2.5$ \\
		
		(\ref{Motion-Predictive-Gain}) & $\kappa = 0.02$ \\
		\hline
	\end{tabular}
\end{table}	

\subsubsection{Implementation Details} 
For given preferred velocity and size ranges of small targets, parameters of the STMD-based neural network are determined by the previous analysis \cite{wang2018directionally}. Parameters $\zeta$, $\eta$ that control the shape of the prediction kernel, are properly tuned based on the preferred velocity range to ensure that the ring area of the prediction kernel can completely cover potential positions of a small target with preferred velocity. Scales of the attention kernel $\Sigma$ are determined by the preferred size range to ensure that small targets with optimal size can obtain the largest response after convolution. Multiple STMDs with different preferred velocity and size ranges could be coordinated to detect small objects with unknown velocities and sizes. Other parameters have been set experimentally, but most remained identical for all test image sequences. The parameter settings for the experimental results are listed in Table \ref{Table-Parameter-pag-STMD}. All experiments are tested on MATLAB software platform under a machine that equips with Intel-i7 2.4 GHz CPU, 16 GB memory.

\subsection{Response Properties of the STMD}
\label{Sec-Response-Properties-of-the-STMD}

\begin{figure}[!t]
	\centering
	\subfloat[]{\includegraphics[width=0.22\textwidth]{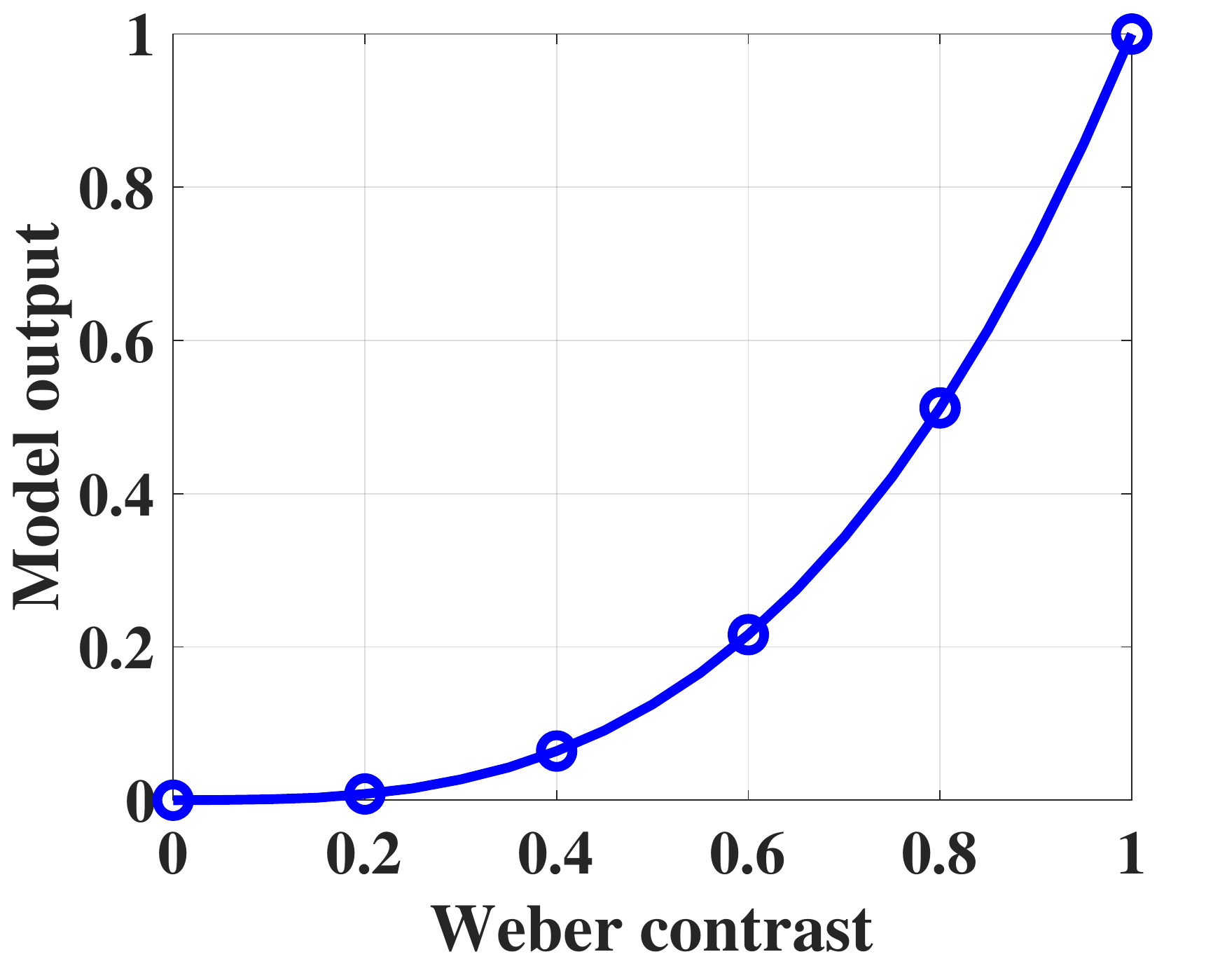}}
	\hfil
	\subfloat[]{\includegraphics[width=0.22\textwidth]{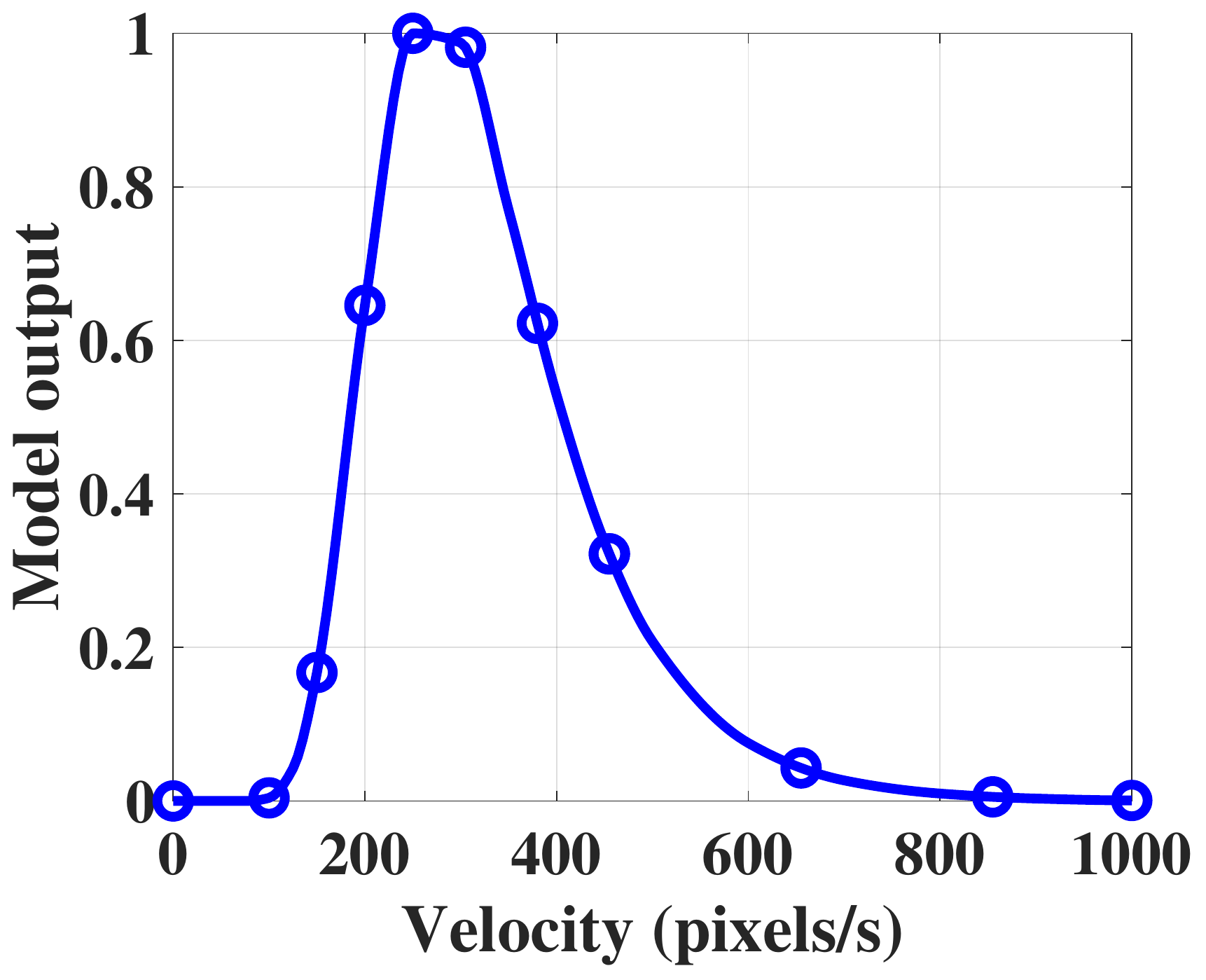}}
	\hfil
	\subfloat[]{\includegraphics[width=0.22\textwidth]{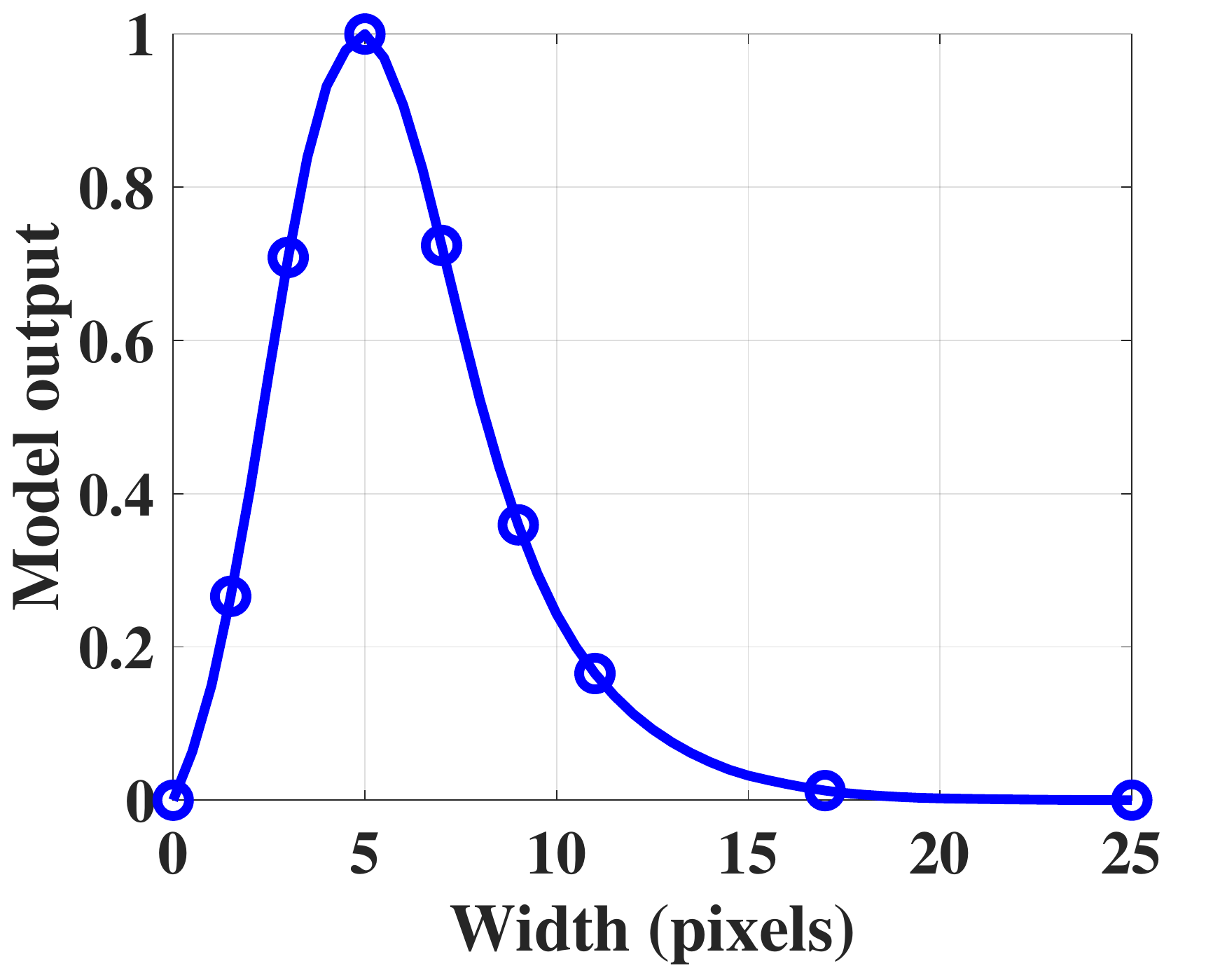}}
	\hfil
	\subfloat[]{\includegraphics[width=0.22\textwidth]{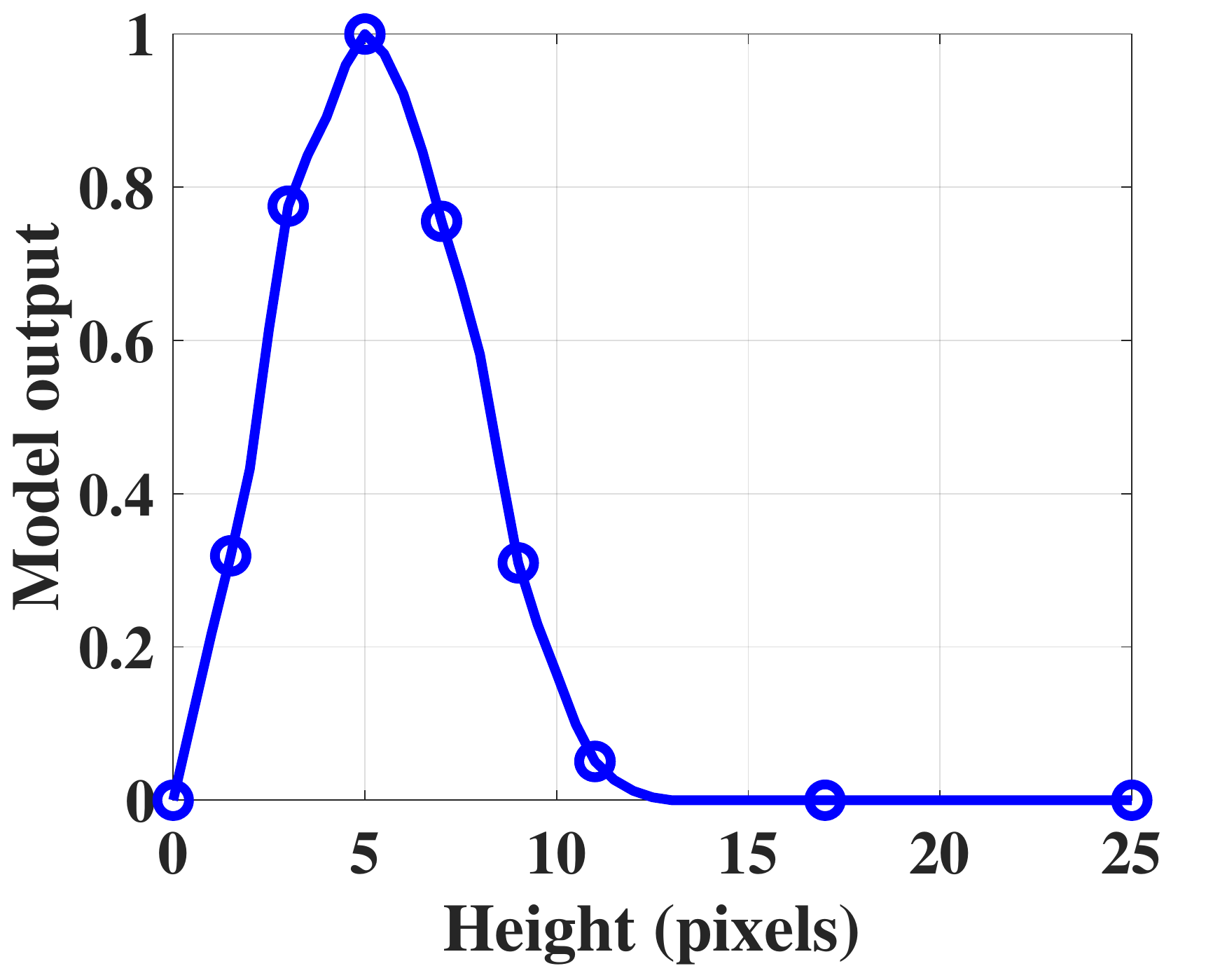}}
	\caption{Outputs of the STMD to a moving object against white background with respect to (a) different Weber contrast, (b) velocities, (c) widths, and (d) heights.}
	\label{Tuning-Properties-pag-STMD}
\end{figure}

To reveal response properties of the STMD-based neural network, we report its outputs $E(x,y,t,\theta)$ to a moving object with different Weber contrast, velocities, widths, and heights. For an object whose size equates to $w \times h$ pixels, we define its neighbouring area as a rectangle with size of  $(w+2d)\times(h+2d)$ pixels where $d$ is set to $10$ pixels \cite{wang2018directionally,wang2019Robust}. Weber contrast can then be given by
\begin{equation}
	\text{Weber contrast} = \frac{|\mu_t - \mu_b|}{255}
	\label{Weber-Contrast-Definition} 
\end{equation}
where $\mu_t$ and $\mu_b$ denote the average pixel intensity of the object and its neighbouring area, respectively. The four parameters of the object, i.e., Weber contrast, velocity, width, and height, are initialized to $1$, $250$ pixels/s, $5$ pixels, and $5$ pixels, respectively. Four experiments are conducted to analyse the STMD outputs with respect to different target parameters,  each of which involves changing one of the parameters while maintaining the other three at their initial values. The recorded outputs to a moving target against white background are shown in Fig. \ref{Tuning-Properties-pag-STMD}.

As observed from Fig. \ref{Tuning-Properties-pag-STMD}(a), the increase in Weber contrast of the object leads to the increase in the STMD output, where the strongest response is reached at Weber contrast $=1$. In Fig. \ref{Tuning-Properties-pag-STMD}(b), we can find that the output of the STMD  is larger than $0$ in the interval $[100,800]$ pixels/s and reaches its maximum at $250$ pixels/s, which correspond to the preferred velocity range and optimal velocity of the STMD, respectively. Fig. \ref{Tuning-Properties-pag-STMD}(c) and (d) presents the outputs of the STMD to objects with different widths and heights. As can be seen, the STMD responds to objects with widths and heights lower than $18$ and $13$ pixels, respectively. In addition, its output peaks at width$=5$ pixels and height$=5$ pixels. 

Fig. \ref{Tuning-Properties-pag-STMD}(a)-(d) provides a good fit to the response properties of the STMD neurons revealed in biological research \cite{nordstrom2006insect,barnett2007retinotopic,nordstrom2012neural}, which means the proposed STMD model displays contrast sensitivity, velocity, width, and height selectivities, respectively.

\subsection{Effectiveness of the Attention Module}

\begin{figure}[t!]
	\centering
	\includegraphics[width=0.25\textwidth]{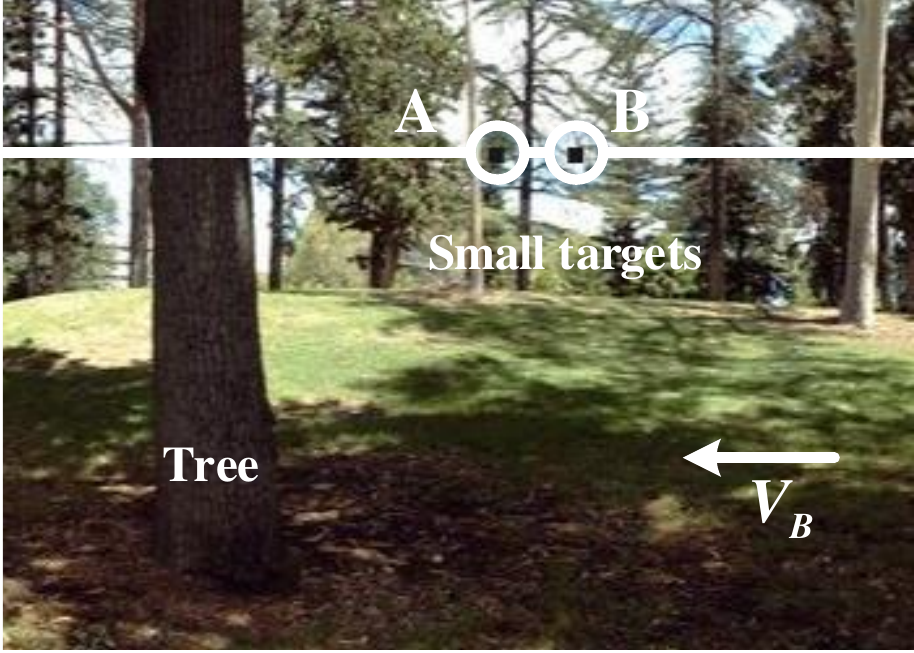}
	\caption{Input image at time $t_0 = 760$ ms where two small targets $A$ and $B$ are moving against the complex background. The background velocity is set as $250$ pixels/s, and arrow $V_B$ denotes its motion direction. The tree is considered as a large object moving with the background at the same velocity.}
	\label{Neural-Output-Input-Image-760}
\end{figure}

\begin{figure}[t!]
	\centering
	\subfloat[]{\includegraphics[width=0.40\textwidth]{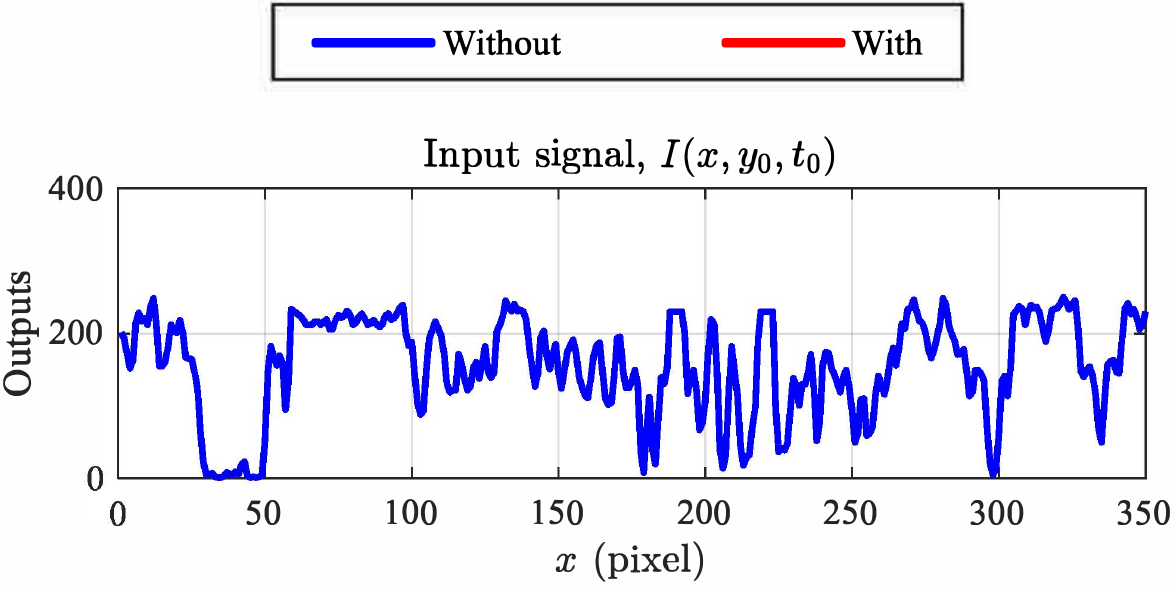}}
	\hfil
	\subfloat[]{\includegraphics[width=0.40\textwidth]{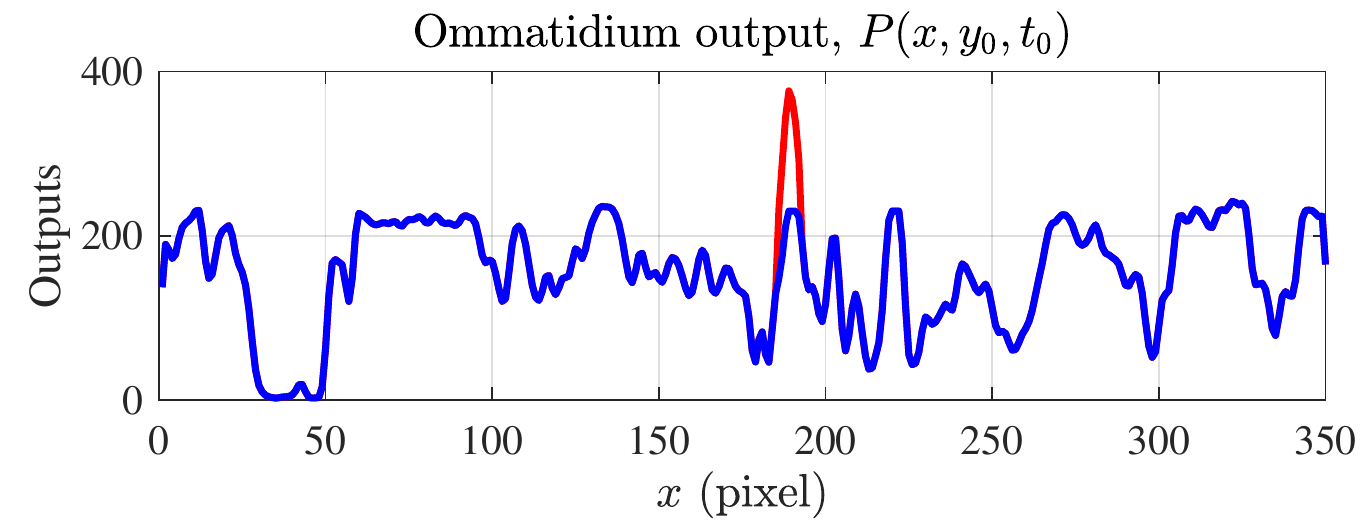}}
	\hfil
	\subfloat[]{\includegraphics[width=0.40\textwidth]{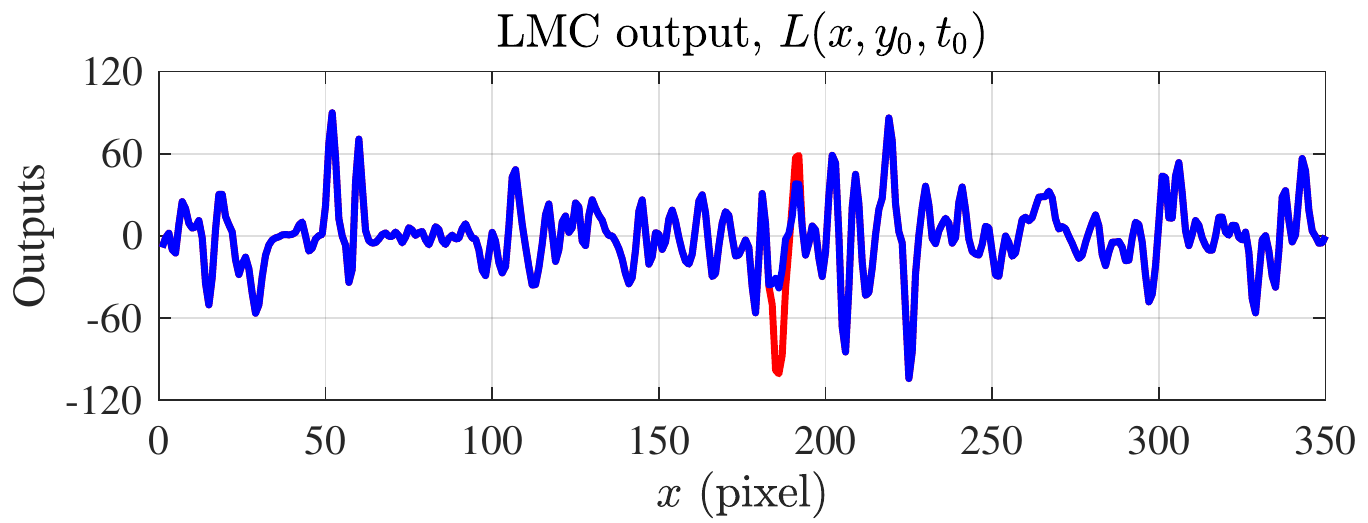}}
	\caption{(a) Input luminance signal $I(x,y_0,t_0)$ where $y_0= 190$ pixels and $t_0=760$ ms. Comparison of (b) Ommatidium outputs $P(x,y_0,t_0)$, (c) LMC output $L(x,y_0,t_0)$ with and without the attention module.}
	\label{Neural-Outputs-1}
\end{figure}

\begin{figure}[t!]
	\centering
	\subfloat[]{\includegraphics[width=0.40\textwidth]{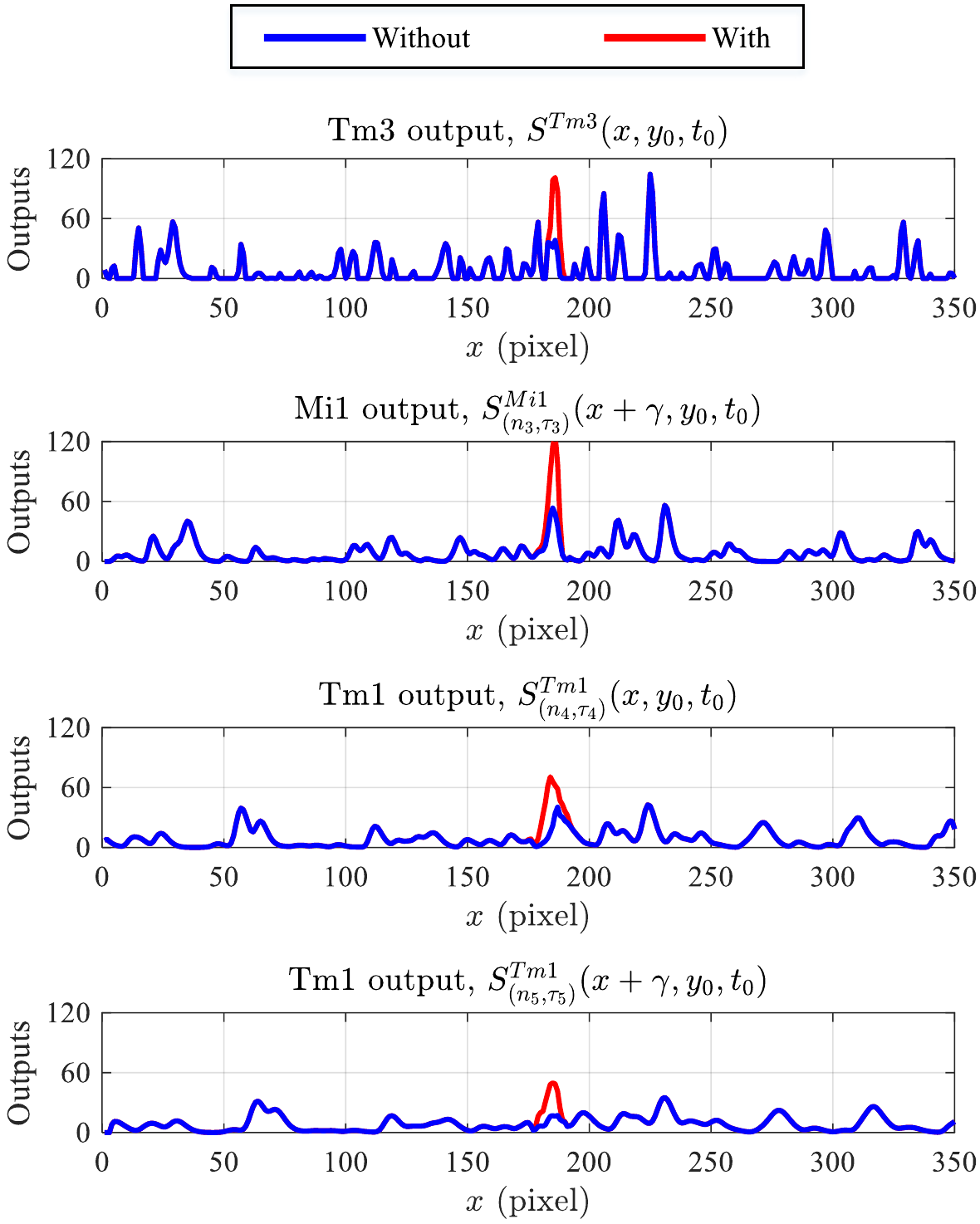}}
	\hfil
	\subfloat[]{\includegraphics[width=0.40\textwidth]{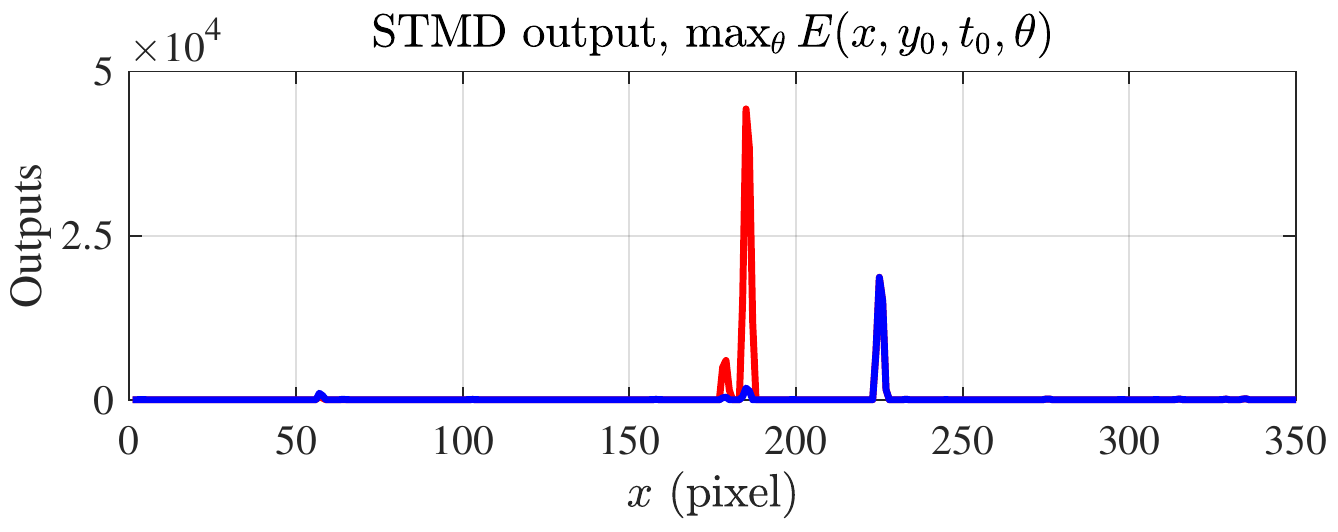}}
	\caption{Comparison of (a) medulla neural outputs, (b) STMD output $E(x,y_0,t_0)$ with and without the attention module.}
	\label{Neural-Outputs-2}
\end{figure}

As described in Section \ref{Section-Attention-Module}, we design an attention module to overcome the heavy dependence of the STMD-based neural network on target contrast against complex background. To validate its effectiveness, we conduct a performance comparison between the STMD-based neural networks with and without an attention module. Fig. \ref{Neural-Output-Input-Image-760} shows the input image $I(x,y,t)$ at time $t = 760$ ms. As can be seen, two small targets are moving against the complex background where the target $A$ shows much lower contrast against its surrounding background compared to the target $B$. In addition, the target $A$ has relative movement to the background, whereas the target $B$ remains static relative to the background. To clearly illustrate signal processing, we observe the input signal $I(x,y_0,t_0)$ with respect to $x$ by setting $y_0 = 190$ pixels in Fig. \ref{Neural-Outputs-1}(a), and then analyse its resulting neural outputs with and without the attention module in Fig. \ref{Neural-Outputs-1}(b), (c) and Fig. \ref{Neural-Outputs-2}.

Fig. \ref{Neural-Outputs-1}(b) shows the outputs of the ommatidium with and without the attention module. As can be seen, the ommatidium smooths the input luminance signal by applying Gaussian blur. The attention module further enhances the contrast of the small target $A$ against surrounding background by adding its convolution output to the smoothed input signal. In comparison, target $B$ does not receive attention, because it remains static relative to the background and is regarded as a part of the background. As shown in Fig. \ref{Neural-Outputs-1}(c), the LMC computes changes of luminance over time for each pixel. Its positive output reflects the increase in luminance while the negative output indicates the decrease in luminance. Since the contrast of the small target $A$ has been strengthened, it induces much more significant luminance changes with respect to time.  

Fig. \ref{Neural-Outputs-2}(a) illustrates four medulla neural outputs with and without the attention module. The medulla neural outputs are derived from either positive or negative components of the output of the LMC, so the effect of the attention module on medulla neurons will be consistent with that on the LMCs in Fig. \ref{Neural-Outputs-1}(c). As can be seen, the medulla neural outputs at the position of the target $A$ are all strengthened by the attention module, whereas the neural outputs at other positions remain unchanged. These four medulla neural outputs that have been aligned in the time domain by time delay and spatial shift, are multiplied together to define the STMD output. Fig. \ref{Neural-Outputs-2}(b) shows the maximal output of the STMD over the direction $\theta$, i.e., $\max_{\theta}E(x,y_0,t_0,\theta)$. It can be observed that the attention module significantly enhances the STMD response to the low-contrast target $A$, while maintaining response to the high-contrast target $B$ that moves relatively static to the background. Note that the STMD-based neural networks with and without attention module all exhibit no response to the large objects, i.e., the tree.

\subsection{Facilitation Effect of the Prediction Module}

Predictive mechanism is able to boost the STMD neural responses, enhance contrast sensitivity and direction selectivity, and facilitate the pursuit of occluded objects, as revealed in biological research \cite{wiederman2017predictive}. To validate the above facilitation effect of the proposed prediction module, we conduct four experiments which are reported in the following subsections.

\subsubsection{Facilitation in Neural Responses}

\begin{figure}[!t]
	\centering
	\includegraphics[width=0.37\textwidth]{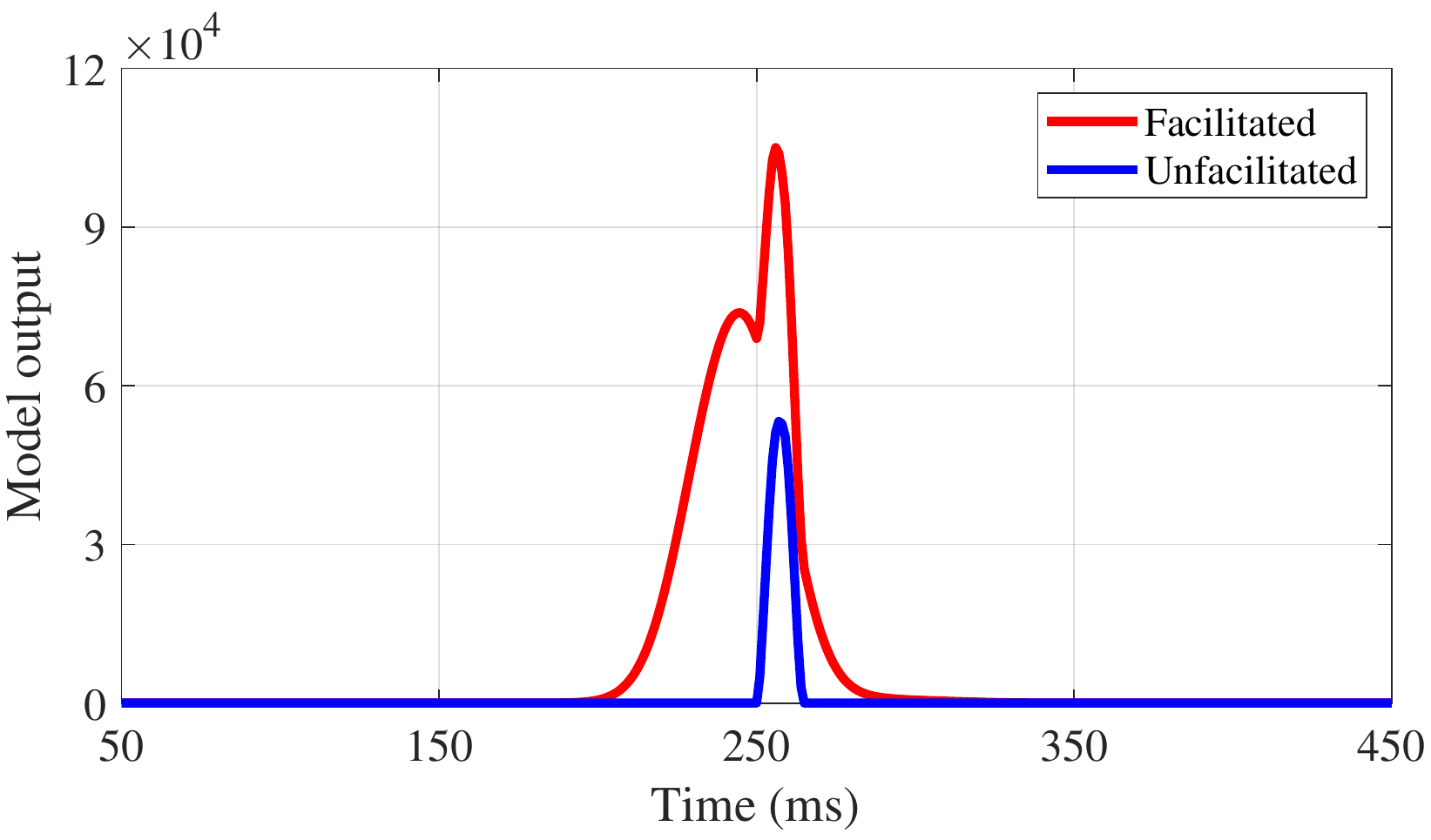}
	\caption{Unfacilitated and facilitated STMD neural responses to a small target at a pixel with respect to time.}
	\label{DSTMD-Output-and-Prediction-Gain-Over-Time}
\end{figure}

\begin{figure}[t!]
	\centering
	\subfloat[]{\includegraphics[width=0.24\textwidth]{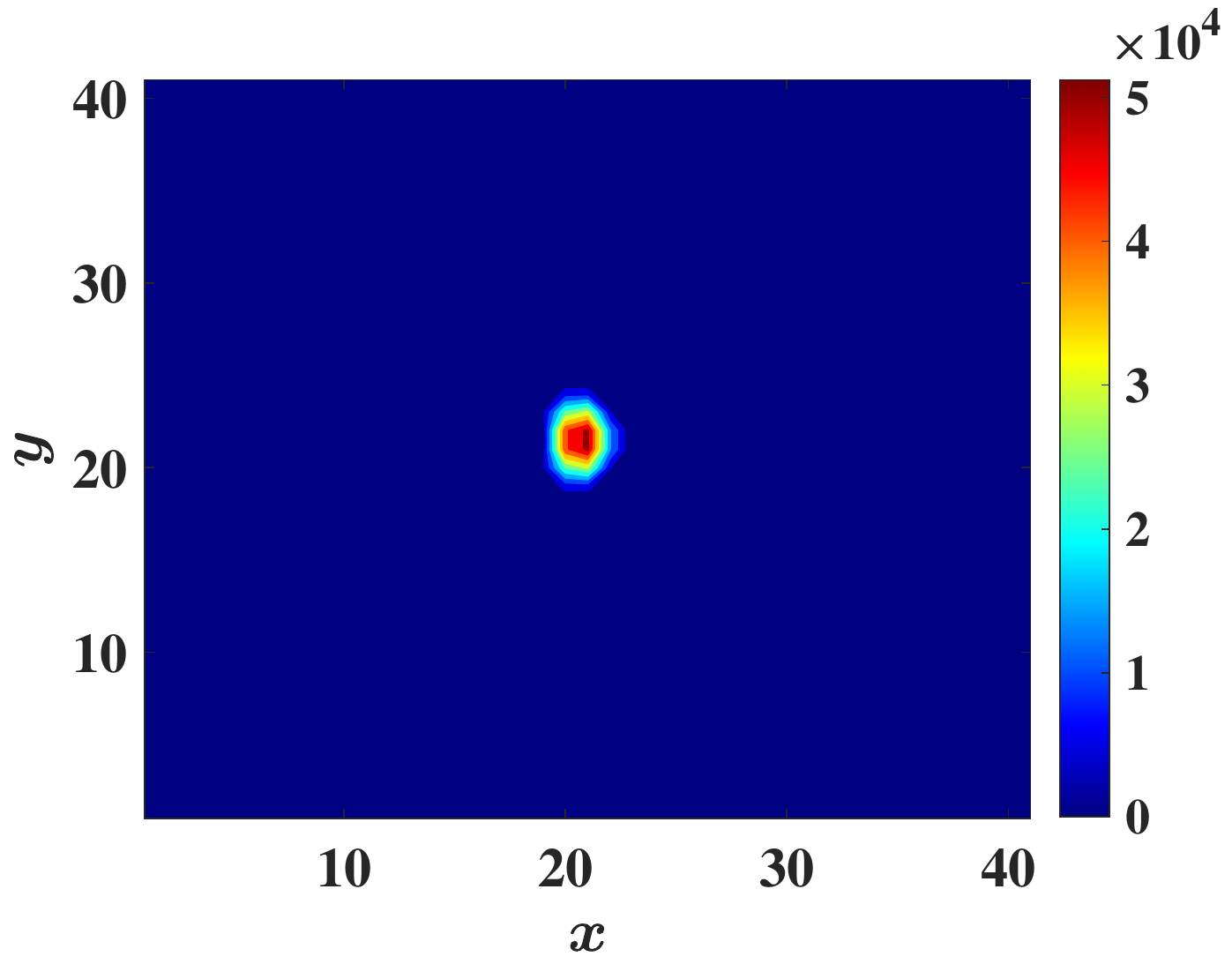}}
	\hfil
	\subfloat[]{\includegraphics[width=0.24\textwidth]{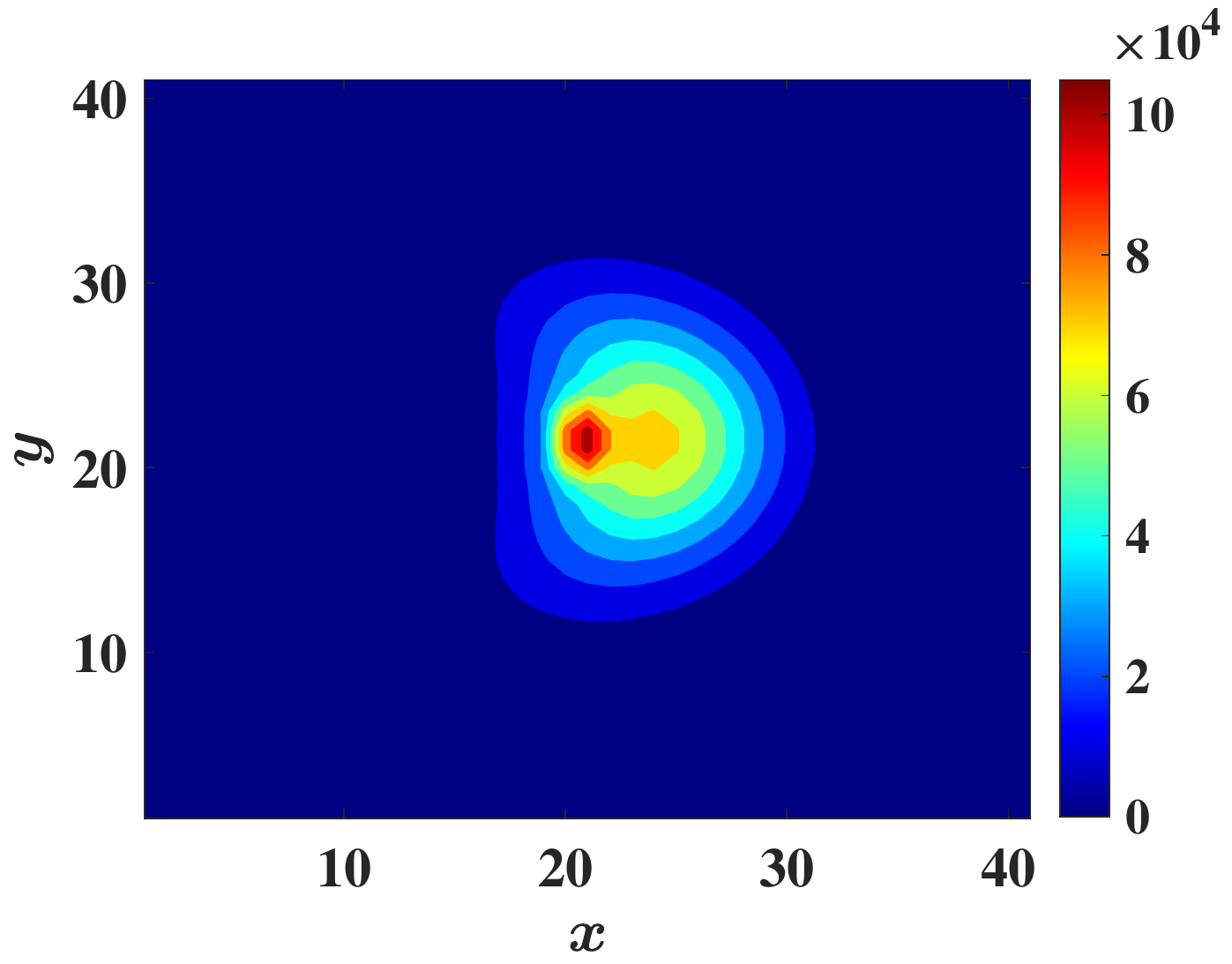}}
	\caption{Planar representations of (a) Unfacilitated and (b) facilitated STMD neural responses to a small target where direction $\theta$ and time $t$ are equal to $\theta = 0$ and $t = 320$ ms, respectively.}
	\label{Planar-Representations-Unfacilitated-Facilitated-Responses}
\end{figure}

We initially compare the STMD neural outputs with and without facilitation where the unfacilitated output $E(x,y,t,\theta)$ and the facilitated output $Q(x,y,t,\theta)$ are defined by (\ref{DSTMD-Lateral-Inhibition}) and (\ref{Motion-Predictive-Gain}), respectively. As depicted in Fig. \ref{DSTMD-Output-and-Prediction-Gain-Over-Time}, the unfacilitated STMD response builds up rapidly to its peak over $7-8$ ms. However, the facilitated response shows a slow build-up lasting roughly $50$ ms before reaching its maximum which is about twice as strong as that of the unfacilitated response. Fig. \ref{Planar-Representations-Unfacilitated-Facilitated-Responses} shows the STMD responses with and without facilitation over the spatial domain. Compared to the unfacilitated response, the prediction module enhances the local STMD responses in a broad region ahead of the target motion direction. 

\subsubsection{Facilitation in Contrast Sensitivity}

\begin{figure}[t]
	\centering
	\subfloat[]{\includegraphics[width=0.18\textwidth]{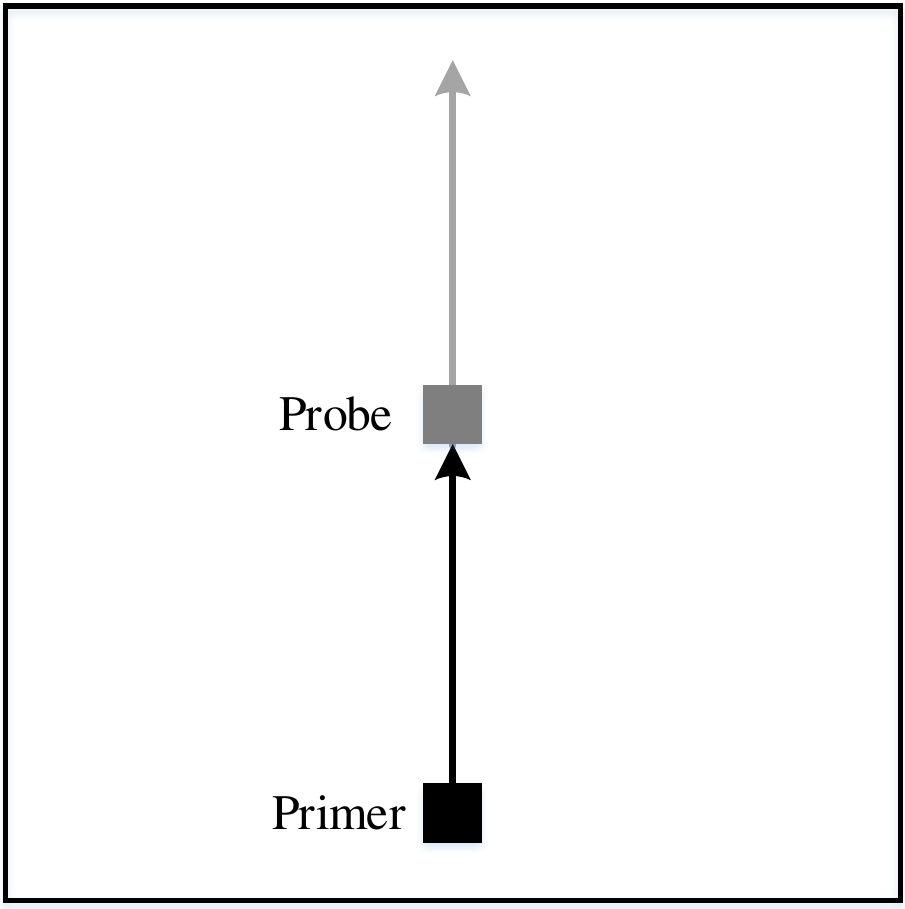}}
	\hfil
	\subfloat[]{\includegraphics[width=0.18\textwidth]{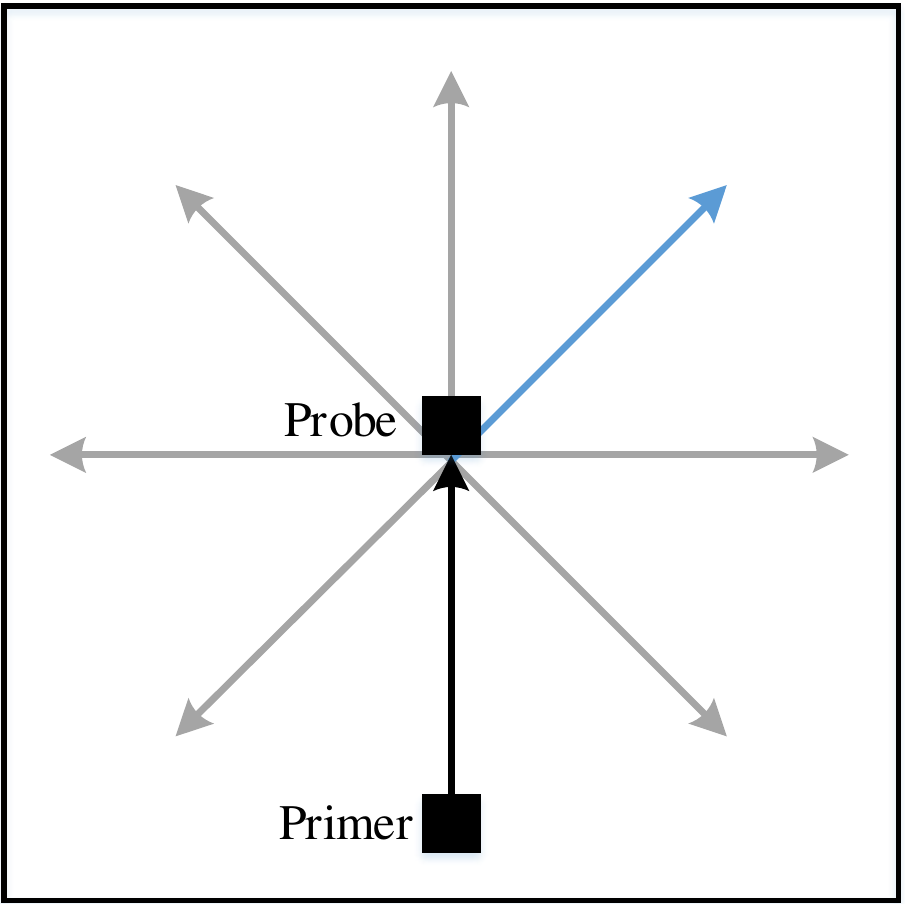}}
	\caption{(a)-(b) Schematics of the Primer \& Probe test for validation of facilitatory effect in contrast sensitivity and direction selectivity, respectively.}
	\label{Schematics-of-the-Primer-and-Probe-test}
\end{figure}

\begin{figure}[t]
	\centering
	\subfloat[]{\includegraphics[width=0.23\textwidth]{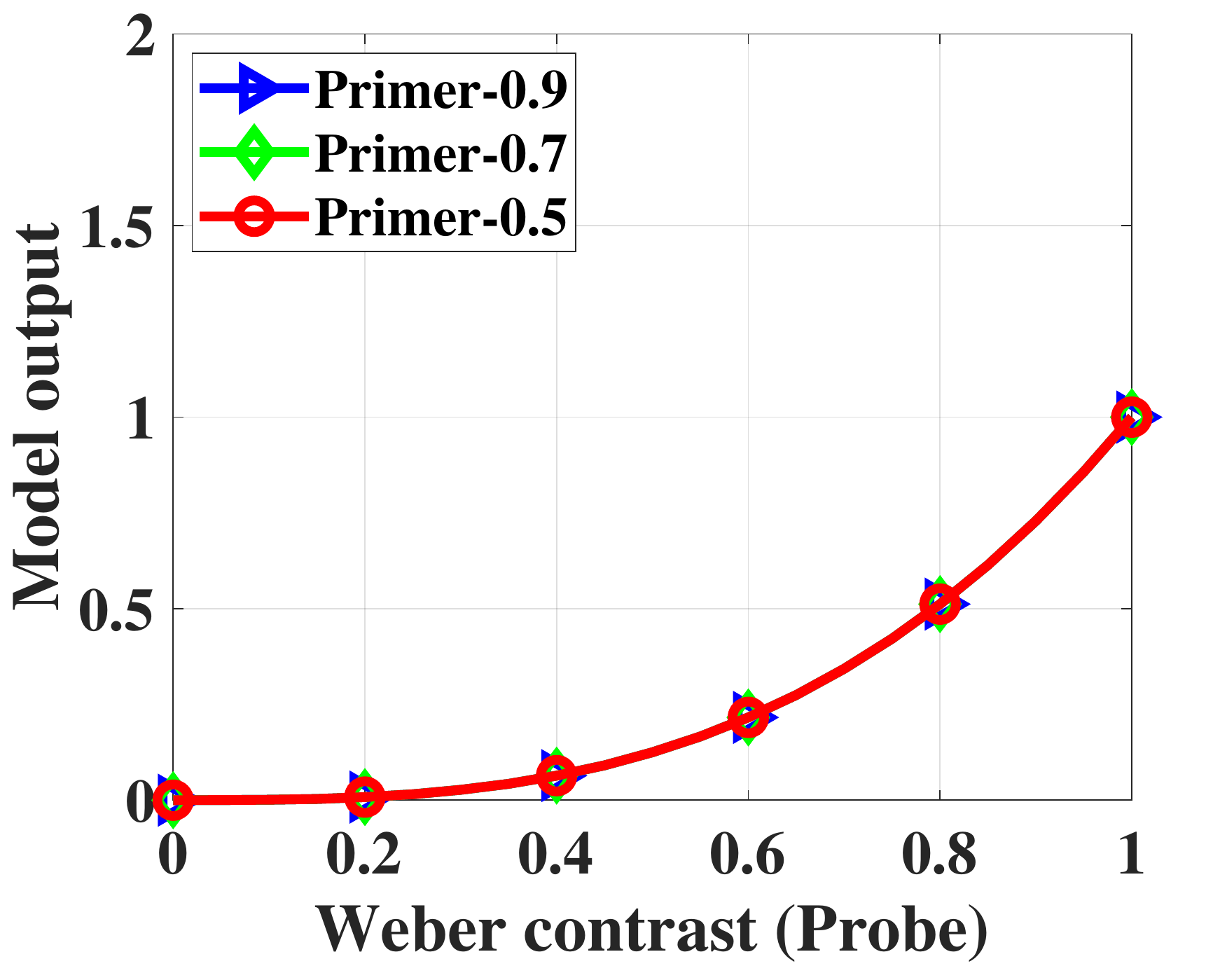}}
	\hfil
	\subfloat[]{\includegraphics[width=0.23\textwidth]{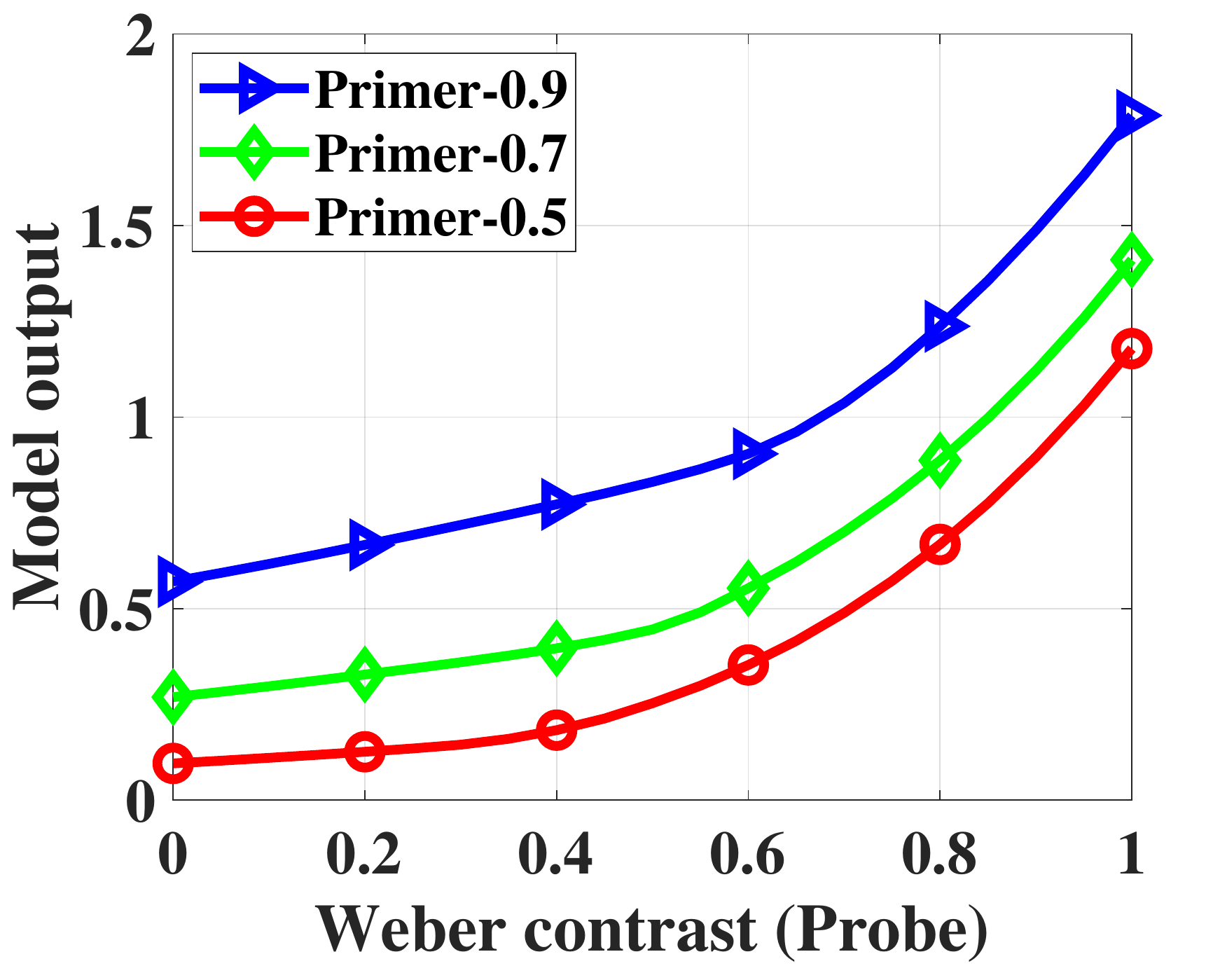}}
	\caption{(a) Unfacilitated and (b) facilitated STMD outputs to the probe with different Weber contrast, preceded by varying-contrast primer.}
	\label{Facilitation-in-Contrast-Sensitivity}
\end{figure}

To validate the facilitatory effect in contrast sensitivity, we conduct a Primer \& Probe experiment as shown in Fig. \ref{Schematics-of-the-Primer-and-Probe-test}(a). Specifically, the input video displays a small target moving along a long path which are divided into two components called the primer and the probe. The primer segment is used to induce spatial facilitation, while the second segment, the probe, is used to record model outputs. The experiment involves changing the Weber contrast of the primer and probe, respectively, and then recording the STMD outputs with and without facilitation to the probe. As depicted in Fig. \ref{Facilitation-in-Contrast-Sensitivity}, the STMD output without facilitation remains unchanged when the contrast of the primer increases. However, for any given contrast of the probe, the STMD output with facilitation shows a significant increase with the growth in the contrast of the primer.

\subsubsection{Facilitation in Direction Selectivity}

\begin{figure}[!t]
	\centering
	\includegraphics[width=0.40\textwidth]{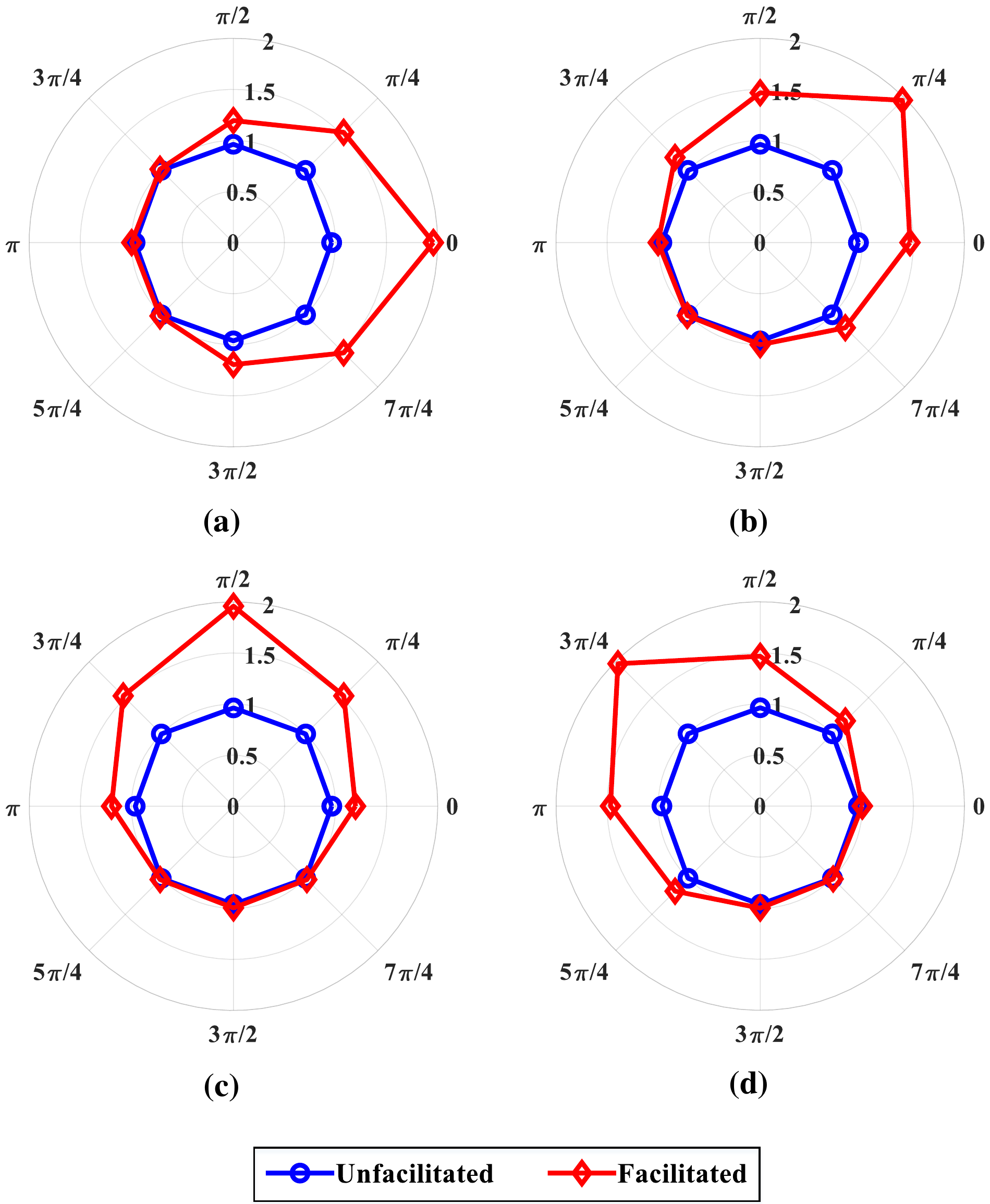}
	\caption{Unfacilitated and facilitated STMD outputs to the probes moving along eight directions $\theta \in \{0, \frac{\pi}{4}, \frac{\pi}{2}, \frac{3\pi}{4}, \pi, \frac{5\pi}{4}, \frac{3\pi}{2}, \frac{7\pi}{4}\}$ when the motion direction of the primer is set to (a) $0$, (b) $\frac{\pi}{4}$, (c) $\frac{\pi}{2}$, (d) $\frac{3\pi}{4}$, respectively.}
	\label{Direction-Facilitation-Primier-Four-Directions}
\end{figure}

To validate the facilitatory effect in direction selectivity, we conduct another Primer \& Probe experiment as shown in Fig. \ref{Schematics-of-the-Primer-and-Probe-test}(b). This involves fixing the Weber contrast of the primer and probe, then changing the angular offset between the primer path and probe path, and finally recording the STMD outputs with and without facilitation to the probe. As can be seen from Fig. \ref{Direction-Facilitation-Primier-Four-Directions}, the motion of the primer shows little effect on the STMD responses without facilitation to the probe. More precisely, the unfacilitated STMD responses along eight directions are equal, regardless of the primer's motion direction. However, the prediction module facilitates the STMD response maximally in the motion direction of the primer. The direction tuning is also shifted to match the motion direction of the primer.

\subsubsection{Facilitation in Pursuit of Occluded Objects}

\begin{figure}[!t]
	\centering
	\includegraphics[width=0.15\textwidth]{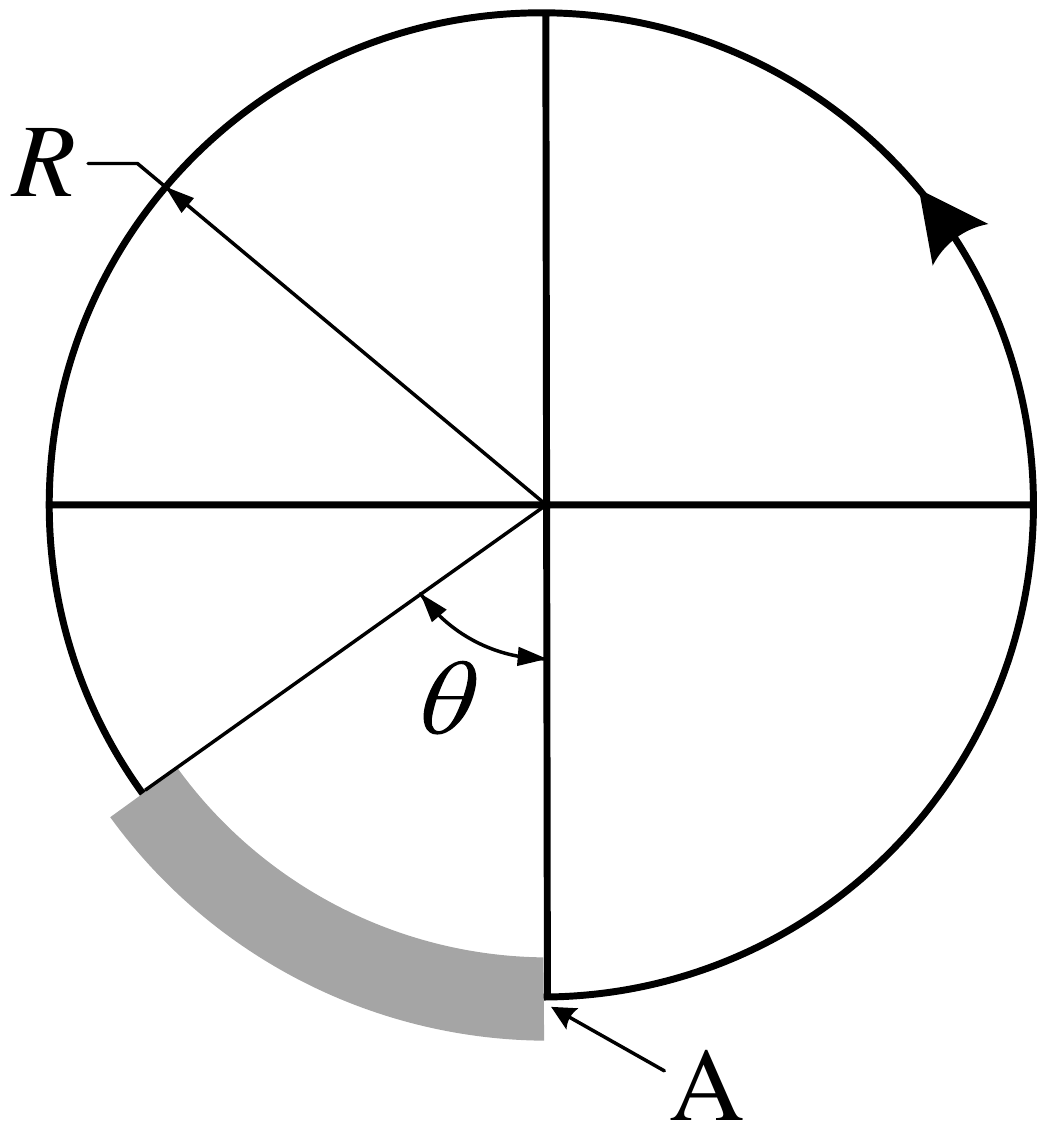}
	\caption{Representation of small target motion in the $xy$ plane where the target is moving counterclockwise along the circular path of radius $R$. During the
		revolution, the target is occluded (grey thick line) where the angle of the occlusion is given by $\theta$ and $A$ denotes the end point of the occlusion.}
	\label{Schematic-Motion-Occlusion}
\end{figure}

\begin{figure}[t!]
	\centering
	\subfloat[]{\includegraphics[width=0.24\textwidth]{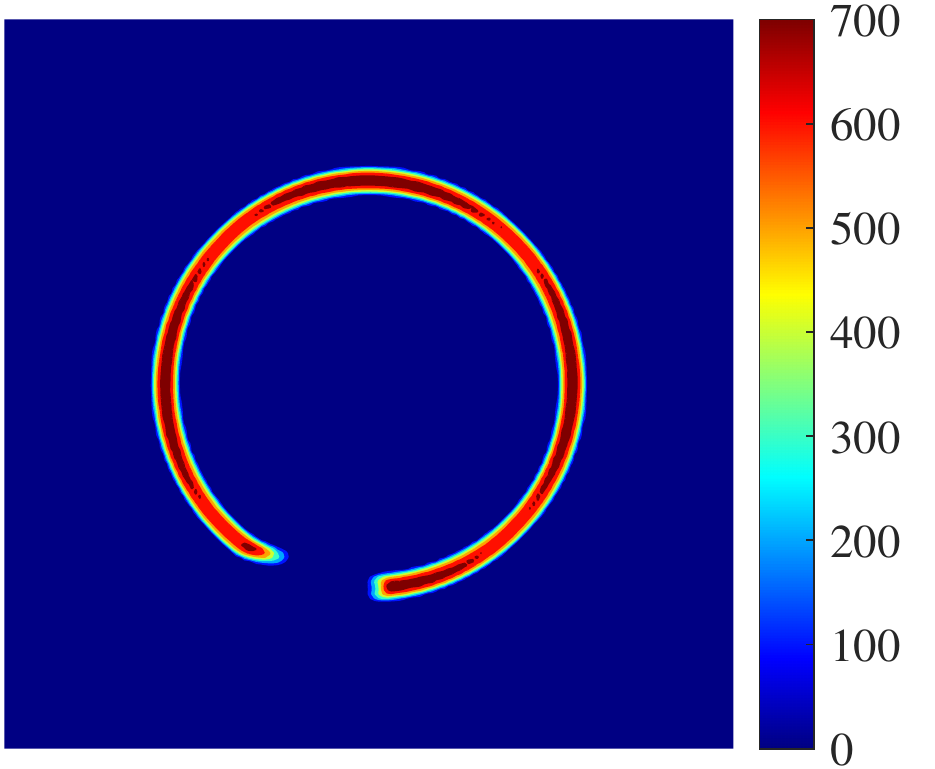}}
	\hfil
	\subfloat[]{\includegraphics[width=0.24\textwidth]{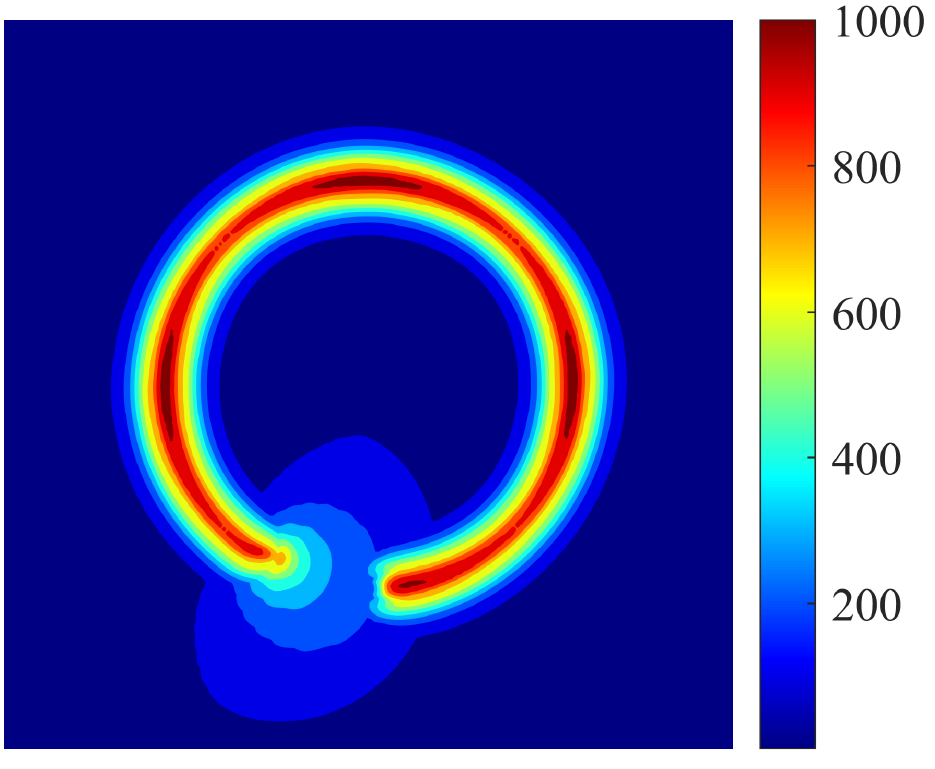}}
	\caption{Planar representations of (a) unfacilitated and (b) facilitated STMD outputs summed on the circular path where the radius $R$ and the occlusion angle $\theta$ is set to $50$ pixels and $30^{\circ}$, respectively. For better visualization, the square root of the output is displayed.}
	\label{Color-Map-DSTMD-Output-Predictive-Gain}
\end{figure}

\begin{figure}[t!]
	\centering
	\subfloat[]{\includegraphics[width=0.24\textwidth]{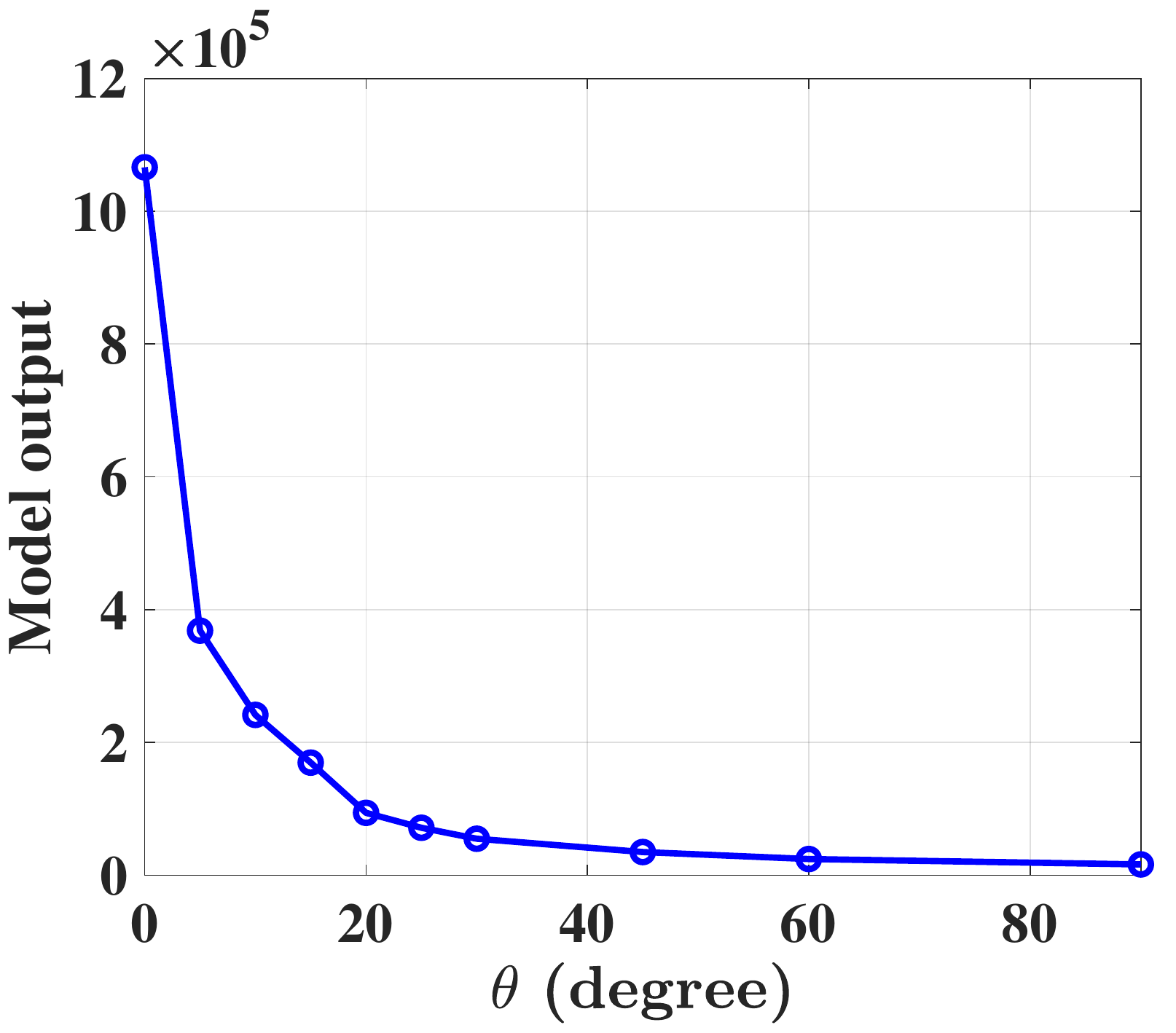}}
	\hfil
	\subfloat[]{\includegraphics[width=0.24\textwidth]{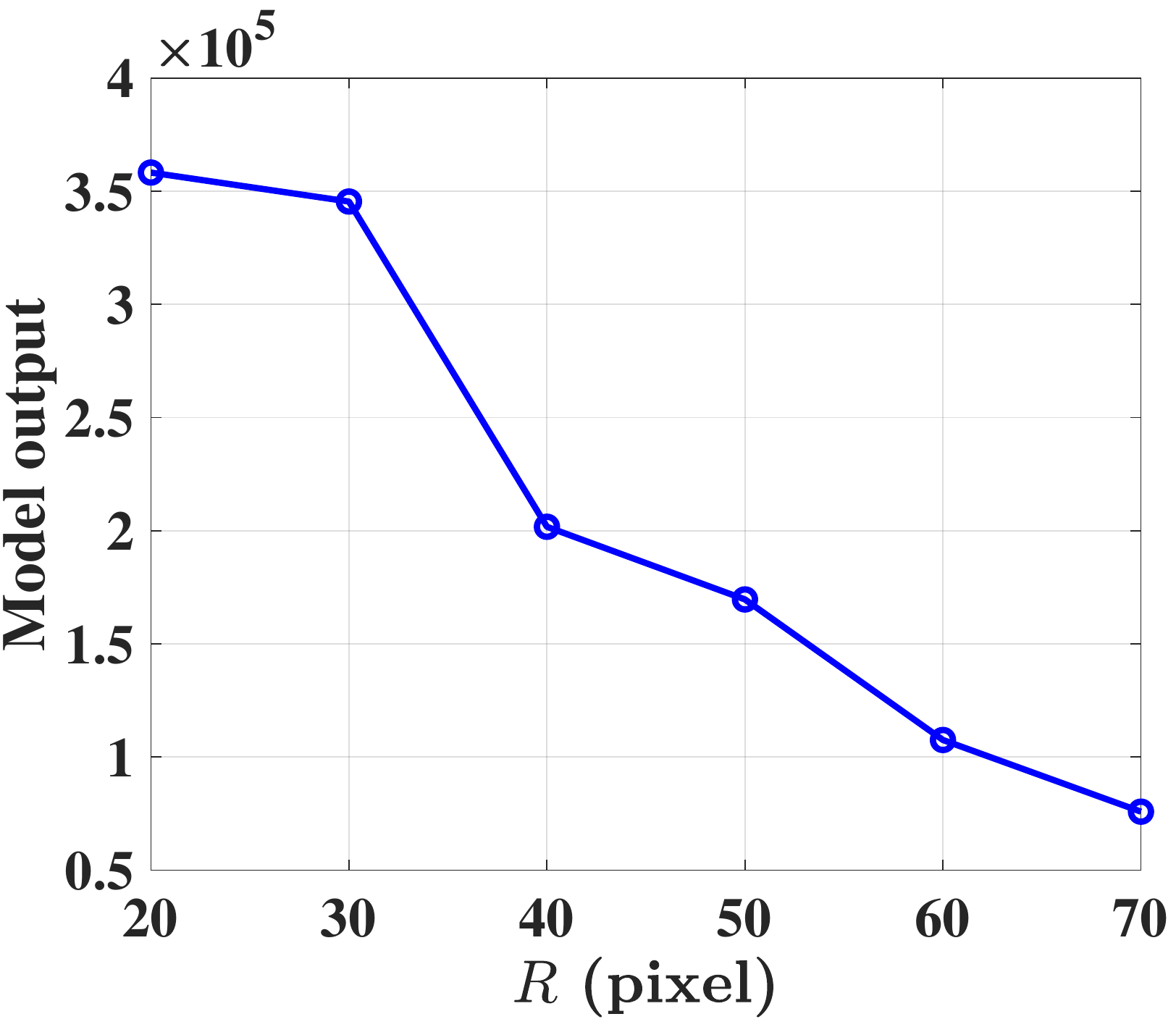}}
	\caption{Outputs of the STMD with facilitation at pixel $A$, i.e., the end point of the occlusion, with respect to (a) occlusion angle $\theta$ and (b) radius $R$.}
	\label{Relationship-STMD-Output-Theta-R}
\end{figure}

To validate the facilitatory effect of the prediction module in smooth pursuit of a moving target that is transiently occluded, we conduct an experiment shown in Fig. \ref{Schematic-Motion-Occlusion}. As can be seen, the input video contains a small target with velocity of $250$ pixels/s moving counterclockwise on a circular path of radius $R$. The target is occluded on part of the circular path where the occlusion angle is set as $\theta$. 

Fig. \ref{Color-Map-DSTMD-Output-Predictive-Gain} shows the summation of the unfacilitated and facilitated STMD outputs along the circular path where the radius $R$ and the occlusion angle $\theta$ are equal to $50$ pixels and $30^{\circ}$, respectively. As can be seen, both the unfacilitated and facilitated STMD responses form a gaped circular path. However, the width of the circle path formed by the facilitated output is much wider than that of the unfacilitated output. In addition,  the facilitated STMD output still spreads forward after the small target disappears. However, the unfacilitated output is close to zero during the occlusion. We further reveal the relationship between the STMD output after the occlusion (pixel $A$) and occlusion angle $\theta$ and radius $R$, respectively. As shown in Fig. \ref{Relationship-STMD-Output-Theta-R}, the increase in occlusion angle $\theta$ and radius $R$ will induce the decrease in the STMD output propagated to pixel $A$, which means that precision of prediction will decrease with the increase in occlusion period.

\subsection{Evaluation on Synthetic and Real-World Data Sets}
\label{Sec-Evaluation-on-Synthetic-and-Real-World-Data-Sets}

We compare the proposed model with three state-of-the-art small target motion detection methods, including DSTMD \cite{wang2018directionally}, ESTMD \cite{wiederman2008model}, and STMD Plus \cite{wang2019Robust}, on the synthetic and real-world data sets in terms of the receiver operating characteristics (ROC) curve. The experimental results are reported in the supplementary material. The results demonstrate that the proposed model has greatly improved detection performance for small targets which exhibit extremely low contrast against cluttered background. However, it fails to detect moving objects without any contrast to backgrounds. In such case, human visual systems are also powerless to deal with object detection tasks.  

In insect' visual system, multiple specialized neural circuits extract various cues simultaneously from complex natural environment, such as color \cite{schnaitmann2020color}, depth information \cite{schwegmann2014depth},  and motion trajectories \cite{gonsek2021paths}. However, the contribution of these visual cues to motion detection and their circuit implementation are still unclear. As future work, multiple visual cues may be combined together to further improve performance of the proposed visual system for small target motion detection.

\section{Conclusion}
\label{Conclusion}

This article proposes an attention and prediction guided visual system to detect small targets in complex natural environments.  To mitigate the heavy dependency on target contrast against the background, the proposed visual system introduces an attention module, an STMD-based neural network, and a prediction module, which are arranged in a recurrent architecture. The attention module is designed to search for potential small targets in predicted areas over the input image and enhance their contrast to neighboring backgrounds. The STMD-based neural network is devised to take the contrast-enhanced image as input and detect small moving targets using both motion information and directional contrast. The prediction module is proposed to anticipate future positions of the detected small targets and generate a prediction for next time step. The proposed visual system significantly improves the performance for small target detection in complex natural environment where small targets always exhibit extremely low contrast. The study provides a robust solution for future autonomous systems to detect small targets timely and react appropriately.


\ifCLASSOPTIONcaptionsoff
  \newpage
\fi


\bibliographystyle{IEEEtran}

\bibliography{IEEEabrv,Reference}

\begin{thebibliography}{10}
\providecommand{\url}[1]{#1}
\csname url@samestyle\endcsname
\providecommand{\newblock}{\relax}
\providecommand{\bibinfo}[2]{#2}
\providecommand{\BIBentrySTDinterwordspacing}{\spaceskip=0pt\relax}
\providecommand{\BIBentryALTinterwordstretchfactor}{4}
\providecommand{\BIBentryALTinterwordspacing}{\spaceskip=\fontdimen2\font plus
\BIBentryALTinterwordstretchfactor\fontdimen3\font minus
  \fontdimen4\font\relax}
\providecommand{\BIBforeignlanguage}[2]{{%
\expandafter\ifx\csname l@#1\endcsname\relax
\typeout{** WARNING: IEEEtran.bst: No hyphenation pattern has been}%
\typeout{** loaded for the language `#1'. Using the pattern for}%
\typeout{** the default language instead.}%
\else
\language=\csname l@#1\endcsname
\fi
#2}}
\providecommand{\BIBdecl}{\relax}
\BIBdecl

\bibitem{qiao2021survey}
H.~Qiao, J.~Chen, and X.~Huang, ``A survey of brain-inspired intelligent
  robots: Integration of vision, decision, motion control, and musculoskeletal
  systems,'' \emph{IEEE Trans. Cybern.}, to be published, doi:
  10.1109/TCYB.2021.3071312.

\bibitem{semnani2020force}
S.~H. Semnani, A.~H.~J. de~Ruiter, and H.~H.~T. Liu, ``Force-based algorithm
  for motion planning of large agent,'' \emph{IEEE Trans. Cybern.}, to be
  published, doi: 10.1109/TCYB.2020.2994122.

\bibitem{yu2019bayesian}
X.~Yu, W.~He, Y.~Li, C.~Xue, J.~Li, J.~Zou, and C.~Yang, ``Bayesian estimation
  of human impedance and motion intention for human-robot collaboration,''
  \emph{IEEE Trans. Cybern.}, vol.~51, no.~4, pp. 1822--1834, Apr. 2019.

\bibitem{Youtube-UAV-Bird}
[Online], Avaliable:
  \url{https://www.youtube.com/watch?v=wphymrmtkjI&list=FL_OHabe8rogCpinac5KHGYA&index=1},
  accessed Aug. 11, 2020.

\bibitem{saleemi2013multiframe}
I.~Saleemi and M.~Shah, ``Multiframe many--many point correspondence for
  vehicle tracking in high density wide area aerial videos,'' \emph{Int. J.
  Comput. Vision}, vol. 104, no.~2, pp. 198--219, Sep. 2013.

\bibitem{javed2018moving}
S.~Javed, A.~Mahmood, S.~Al-Maadeed, T.~Bouwmans, and S.~K. Jung, ``Moving
  object detection in complex scene using spatiotemporal structured-sparse
  rpca,'' \emph{IEEE Trans. Image Process.}, vol.~28, no.~2, pp. 1007--1022,
  Oct. 2018.

\bibitem{fortun2015optical}
D.~Fortun, P.~Bouthemy, and C.~Kervrann, ``Optical flow modeling and
  computation: a survey,'' \emph{Comput. Vis. Image Underst.}, vol. 134, pp.
  1--21, May 2015.

\bibitem{Redmon_2016_CVPR}
J.~Redmon, S.~Divvala, R.~Girshick, and A.~Farhadi, ``You only look once:
  Unified, real-time object detection,'' in \emph{Proc. IEEE Conf. Comput. Vis.
  Pattern Recognit.}, June 2016, pp. 779--788.

\bibitem{tang2017multiview}
J.~Tang, Y.~Tian, P.~Zhang, and X.~Liu, ``Multiview privileged support vector
  machines,'' \emph{IEEE Trans. Neural Netw. Learn. Syst.}, vol.~29, no.~8, pp.
  3463--3477, Aug. 2017.

\bibitem{zhang2011evolutionary}
J.~Zhang, Z.-h. Zhan, Y.~Lin, N.~Chen, Y.-j. Gong, J.-h. Zhong, H.~S. Chung,
  Y.~Li, and Y.-h. Shi, ``Evolutionary computation meets machine learning: A
  survey,'' \emph{IEEE Computational Intelligence Magazine}, vol.~6, no.~4, pp.
  68--75, Oct. 2011.

\bibitem{wang2020bioinspired}
H.~Wang, Q.~Fu, H.~Wang, P.~Baxter, J.~Peng, and S.~Yue, ``A bioinspired
  angular velocity decoding neural network model for visually guided flights,''
  \emph{Neural Netw.}, vol. 136, pp. 180--193, Apr. 2021.

\bibitem{sun2020decentralised}
X.~Sun, S.~Yue, and M.~Mangan, ``A decentralised neural model explaining
  optimal integration of navigational strategies in insects,'' \emph{Elife},
  vol.~9, p. e54026, 2020.

\bibitem{fu2019towards}
Q.~Fu, H.~Wang, C.~Hu, and S.~Yue, ``Towards computational models and
  applications of insect visual systems for motion perception: A review,''
  \emph{Artif. Life}, vol.~25, no.~3, pp. 263--311, 2019.

\bibitem{nordstrom2006insect}
K.~Nordstr{\"o}m, P.~D. Barnett, and D.~C. O'Carroll, ``Insect detection of
  small targets moving in visual clutter,'' \emph{PLoS Biol.}, vol.~4, no.~3,
  p. e54, Feb. 2006.

\bibitem{barnett2007retinotopic}
P.~D. Barnett, K.~Nordstr{\"o}m, and D.~C. O'Carroll, ``Retinotopic
  organization of small-field-target-detecting neurons in the insect visual
  system,'' \emph{Curr. Biol.}, vol.~17, no.~7, pp. 569--578, Apr. 2007.

\bibitem{nordstrom2012neural}
K.~Nordstr{\"o}m, ``Neural specializations for small target detection in
  insects,'' \emph{Curr. Opin. Neurobiol.}, vol.~22, no.~2, pp. 272--278, Apr.
  2012.

\bibitem{wiederman2008model}
S.~D. Wiederman, P.~A. Shoemaker, and D.~C. O'Carroll, ``A model for the
  detection of moving targets in visual clutter inspired by insect
  physiology,'' \emph{PLoS One}, vol.~3, no.~7, pp. 1--11, Jul. 2008.

\bibitem{wiederman2013biologically}
S.~D. Wiederman and D.~C. O’Carroll, ``Biologically inspired feature
  detection using cascaded correlations of off and on channels,'' \emph{J.
  Artif. Intell. Soft Comput. Res.}, vol.~3, no.~1, pp. 5--14, Dec. 2013.

\bibitem{wang2018directionally}
H.~Wang, J.~Peng, and S.~Yue, ``A directionally selective small target motion
  detecting visual neural network in cluttered backgrounds,'' \emph{IEEE Trans.
  Cybern.}, vol.~50, no.~4, pp. 1541--1555, Apr. 2020.

\bibitem{wang2019Robust}
H.~Wang, J.~Peng, X.~Zheng, and S.~Yue, ``A robust visual system for small
  target motion detection against cluttered moving backgrounds,'' \emph{IEEE
  Trans. Neural Netw. Learn. Syst.}, vol.~31, no.~3, pp. 839--853, Mar. 2020.

\bibitem{nityananda2016attention}
V.~Nityananda, ``Attention-like processes in insects,'' \emph{Proc. R. Soc. B},
  vol. 283, no. 1842, p. 20161986, Nov. 2016.

\bibitem{schroger2015bridging}
E.~Schr{\"o}ger, S.~A. Kotz, and I.~SanMiguel, ``Bridging prediction and
  attention in current research on perception and action,'' \emph{Brain
  Research}, vol. 1626, pp. 1--13, Nov. 2015.

\bibitem{bagheri2020evidence}
Z.~M. Bagheri, C.~G. Donohue, and J.~M. Hemmi, ``Evidence of predictive
  selective attention in fiddler crabs during escape in the natural
  environment,'' \emph{J. Exp. Biol}, vol. 223, no.~21, Nov. 2020.

\bibitem{wiederman2017predictive}
S.~D. Wiederman, J.~M. Fabian, J.~R. Dunbier, and D.~C. O’Carroll, ``A
  predictive focus of gain modulation encodes target trajectories in insect
  vision,'' \emph{Elife}, vol.~6, p. e26478, Jul. 2017.

\bibitem{rind2016two}
F.~C. Rind, S.~Wernitznig, P.~P{\"o}lt, A.~Zankel, D.~G{\"u}tl, J.~Sztarker,
  and G.~Leitinger, ``Two identified looming detectors in the locust:
  ubiquitous lateral connections among their inputs contribute to selective
  responses to looming objects,'' \emph{Sci. Rep.}, vol.~6, no.~1, pp. 1--16,
  Oct. 2016.

\bibitem{rind1996neural}
F.~C. Rind and D.~Bramwell, ``Neural network based on the input organization of
  an identified neuron signaling impending collision,'' \emph{J. Neurophysiol},
  vol.~75, no.~3, pp. 967--985, Mar. 1996.

\bibitem{maisak2013directional}
M.~S. Maisak, J.~Haag, G.~Ammer, E.~Serbe, M.~Meier, A.~Leonhardt,
  T.~Schilling, A.~Bahl, G.~M. Rubin, A.~Nern \emph{et~al.}, ``A directional
  tuning map of drosophila elementary motion detectors,'' \emph{Nature}, vol.
  500, no. 7461, pp. 212--216, Aug. 2013.

\bibitem{perry2017generation}
M.~Perry, N.~Konstantinides, F.~Pinto-Teixeira, and C.~Desplan, ``Generation
  and evolution of neural cell types and circuits: insights from the drosophila
  visual system,'' \emph{Annual review of genetics}, vol.~51, pp. 501--527,
  2017.

\bibitem{yue2006collision}
S.~Yue and F.~C. Rind, ``Collision detection in complex dynamic scenes using an
  lgmd-based visual neural network with feature enhancement,'' \emph{IEEE
  Trans. Neural Netw.}, vol.~17, no.~3, pp. 705--716, May 2006.

\bibitem{yue2013redundant}
------, ``Redundant neural vision systems-competing for collision recognition
  roles,'' \emph{IEEE Trans. Auton. Mental Develop.}, vol.~5, no.~2, pp.
  173--186, Apr. 2013.

\bibitem{hu2016bio}
C.~Hu, F.~Arvin, C.~Xiong, and S.~Yue, ``Bio-inspired embedded vision system
  for autonomous micro-robots: the lgmd case,'' \emph{IEEE Trans. Cogn.
  Develop. Syst.}, vol.~9, no.~3, pp. 241--254, Sep. 2016.

\bibitem{fu2019robust}
Q.~Fu, C.~Hu, J.~Peng, F.~C. Rind, and S.~Yue, ``A robust collision perception
  visual neural network with specific selectivity to darker objects,''
  \emph{IEEE Trans. Cybern.}, vol.~50, no.~12, pp. 5074--5088, Dec. 2019.

\bibitem{zhao2021enhancing}
J.~Zhao, H.~Wang, N.~Bellotto, C.~Hu, J.~Peng, and S.~Yue, ``Enhancing lgmd’s
  looming selectivity for uav with spatial–temporal distributed presynaptic
  connections,'' \emph{IEEE Trans. Neural Netw. Learn. Syst.}, to be published,
  doi: 10.1109/TNNLS.2021.3106946.

\bibitem{salt2019parameter}
L.~Salt, D.~Howard, G.~Indiveri, and Y.~Sandamirskaya, ``Parameter optimization
  and learning in a spiking neural network for uav obstacle avoidance targeting
  neuromorphic processors,'' \emph{IEEE Trans. Neural Netw. Learn. Syst.},
  vol.~31, no.~9, pp. 3305--3318, Oct. 2019.

\bibitem{eichner2011internal}
H.~Eichner, M.~Joesch, B.~Schnell, D.~F. Reiff, and A.~Borst, ``Internal
  structure of the fly elementary motion detector,'' \emph{Neuron}, vol.~70,
  no.~6, pp. 1155--1164, Jun. 2011.

\bibitem{cope2016model}
A.~J. Cope, C.~Sabo, K.~Gurney, E.~Vasilaki, and J.~A. Marshall, ``A model for
  an angular velocity-tuned motion detector accounting for deviations in the
  corridor-centering response of the bee,'' \emph{PLoS Comput. Biol.}, vol.~12,
  no.~5, p. e1004887, 2016.

\bibitem{bertrand2015bio}
O.~J. Bertrand, J.~P. Lindemann, and M.~Egelhaaf, ``A bio-inspired collision
  avoidance model based on spatial information derived from motion detectors
  leads to common routes,'' \emph{PLoS Comput. Biol.}, vol.~11, no.~11, p.
  e1004339, 2015.

\bibitem{missler1995neural}
J.~M. Missler and F.~A. Kamangar, ``A neural network for pursuit tracking
  inspired by the fly visual system,'' \emph{Neural Netw.}, vol.~8, no.~3, pp.
  463--480, 1995.

\bibitem{nityananda2013bumblebee}
V.~Nityananda and J.~G. Pattrick, ``Bumblebee visual search for multiple
  learned target types,'' \emph{J. Exp. Biol}, vol. 216, no.~22, pp.
  4154--4160, Oct. 2013.

\bibitem{sareen2011attracting}
P.~Sareen, R.~Wolf, and M.~Heisenberg, ``Attracting the attention of a fly,''
  \emph{Proc. Natl. Acad. Sci. U.S.A.}, vol. 108, no.~17, pp. 7230--7235, Apr.
  2011.

\bibitem{mnih2014recurrent}
V.~Mnih, N.~Heess, A.~Graves \emph{et~al.}, ``Recurrent models of visual
  attention,'' \emph{Advances in neural information processing systems},
  vol.~27, pp. 2204--2212, 2014.

\bibitem{yang2016stacked}
Z.~Yang, X.~He, J.~Gao, L.~Deng, and A.~Smola, ``Stacked attention networks for
  image question answering,'' in \emph{Proceedings of the IEEE conference on
  computer vision and pattern recognition}, 2016, pp. 21--29.

\bibitem{vaswani2017attention}
A.~Vaswani, N.~Shazeer, N.~Parmar, J.~Uszkoreit, L.~Jones, A.~N. Gomez,
  {\L}.~Kaiser, and I.~Polosukhin, ``Attention is all you need,'' in
  \emph{Advances in neural information processing systems}, 2017, pp.
  5998--6008.

\bibitem{chen2017sca}
L.~Chen, H.~Zhang, J.~Xiao, L.~Nie, J.~Shao, W.~Liu, and T.-S. Chua, ``Sca-cnn:
  Spatial and channel-wise attention in convolutional networks for image
  captioning,'' in \emph{Proceedings of the IEEE conference on computer vision
  and pattern recognition}, 2017, pp. 5659--5667.

\bibitem{mischiati2015internal}
M.~Mischiati, H.-T. Lin, P.~Herold, E.~Imler, R.~Olberg, and A.~Leonardo,
  ``Internal models direct dragonfly interception steering,'' \emph{Nature},
  vol. 517, no. 7534, pp. 333--338, Dec. 2015.

\bibitem{kooij2019context}
J.~F. Kooij, F.~Flohr, E.~A. Pool, and D.~M. Gavrila, ``Context-based path
  prediction for targets with switching dynamics,'' \emph{Int. J. Comput.
  Vis.}, vol. 127, no.~3, pp. 239--262, 2019.

\bibitem{luber2010people}
M.~Luber, J.~A. Stork, G.~D. Tipaldi, and K.~O. Arras, ``People tracking with
  human motion predictions from social forces,'' in \emph{IEEE Int. Conf.
  Robot. Autom. (ICRA)}.\hskip 1em plus 0.5em minus 0.4em\relax IEEE, 2010, pp.
  464--469.

\bibitem{stulp2015facilitating}
F.~Stulp, J.~Grizou, B.~Busch, and M.~Lopes, ``Facilitating intention
  prediction for humans by optimizing robot motions,'' in \emph{IEEE/RSJ
  international conference on intelligent robots and systems (IROS)}.\hskip 1em
  plus 0.5em minus 0.4em\relax IEEE, 2015, pp. 1249--1255.

\bibitem{behnia2014processing}
R.~Behnia, D.~A. Clark, A.~G. Carter, T.~R. Clandinin, and C.~Desplan,
  ``Processing properties of on and off pathways for drosophila motion
  detection,'' \emph{Nature}, vol. 512, no. 7515, p. 427, Aug. 2014.

\bibitem{warrant2017remarkable}
E.~J. Warrant, ``The remarkable visual capacities of nocturnal insects: vision
  at the limits with small eyes and tiny brains,'' \emph{Philos. Trans. R. Soc.
  B}, vol. 372, no. 1717, p. 20160063, Apr. 2017.

\bibitem{freifeld2013gabaergic}
L.~Freifeld, D.~A. Clark, M.~J. Schnitzer, M.~A. Horowitz, and T.~R. Clandinin,
  ``Gabaergic lateral interactions tune the early stages of visual processing
  in drosophila,'' \emph{Neuron}, vol.~78, no.~6, pp. 1075--1089, Jun. 2013.

\bibitem{takemura2013visual}
S.-y. Takemura, A.~Bharioke, Z.~Lu, A.~Nern, S.~Vitaladevuni, P.~K. Rivlin,
  W.~T. Katz, D.~J. Olbris, S.~M. Plaza, P.~Winston \emph{et~al.}, ``A visual
  motion detection circuit suggested by drosophila connectomics,''
  \emph{Nature}, vol. 500, no. 7461, p. 175, Aug. 2013.

\bibitem{caves2018visual}
E.~M. Caves, N.~C. Brandley, and S.~Johnsen, ``Visual acuity and the evolution
  of signals,'' \emph{Trends Ecol. Evol.}, vol.~33, no.~5, pp. 358--372, May
  2018.

\bibitem{Wang_2017_ICCV}
G.~Wang, C.~Lopez-Molina, and B.~De~Baets, ``Blob reconstruction using
  unilateral second order gaussian kernels with application to high-iso
  long-exposure image denoising,'' in \emph{Proceedings of the IEEE
  International Conference on Computer Vision (ICCV)}, Oct. 2017.

\bibitem{de1991theory}
B.~De~Vries and J.~C. Pr{\'\i}ncipe, ``A theory for neural networks with time
  delays,'' in \emph{Proc. NIPS}, 1990, pp. 162--168.

\bibitem{straw2008vision}
A.~D. Straw, ``Vision egg: an open-source library for realtime visual stimulus
  generation.'' \emph{Front. Neuroinf.}, vol.~2, no.~4, Nov. 2008.

\bibitem{RIST-Data-Set}
\emph{RIST Data Set.} [Online], Avaliable:
  \url{https://sites.google.com/view/hongxinwang-personalsite/download},
  accessed Apr. 6, 2020.

\bibitem{schnaitmann2020color}
C.~Schnaitmann, M.~Pagni, and D.~F. Reiff, ``Color vision in insects: insights
  from drosophila,'' \emph{J. Comp. Physiol. A}, vol. 206, no.~2, pp. 183--198,
  Feb. 2020.

\bibitem{schwegmann2014depth}
A.~Schwegmann, J.~P. Lindemann, and M.~Egelhaaf, ``Depth information in natural
  environments derived from optic flow by insect motion detection system: a
  model analysis,'' \emph{Front. Comput. Neurosci.}, vol.~8, p.~83, Aug. 2014.

\bibitem{gonsek2021paths}
A.~Gonsek, M.~Jeschke, S.~R{\"o}nnau, and O.~J. Bertrand, ``From paths to
  routes: A method for path classification,'' \emph{Front. Behav. Neurosci.},
  vol.~14, p. 274, Jan. 2021.

\end{thebibliography}

\end{document}